\documentclass[preprint,12pt,times, authoryear]{elsarticle}

\usepackage{hyperref}
\usepackage{url}            
\usepackage{booktabs}       
\usepackage{amsfonts}       
\usepackage{nicefrac}       
\usepackage{microtype}      
\usepackage{lipsum}
\usepackage{graphicx}
\usepackage{tabularx}
\usepackage{amsmath}
\usepackage{array}
\graphicspath{ {./images/} }
\usepackage{multirow}
\usepackage{bbding}
\usepackage{algorithm}
\usepackage{algpseudocode}

\usepackage{linegoal}

\newcommand*{\LongState}[1]{\State
\parbox[t]{\linegoal}{#1\strut}}

\usepackage{amssymb}

\journal{Computer, Speech and Language}

\begin{document}

\begin{frontmatter}



\title{An Automated Quality Evaluation Framework of Psychotherapy Conversations with Local Quality Estimates}

\author[add1]{Zhuohao Chen\corref{cor1}}
\cortext[cor1]{Corresponding author:
Tel.: +1-213-740-4146;  
fax: +1-213-740-4651;
}
\ead{zhuohaoc@usc.edu}
\author[add1]{Nikolaos Flemotomos}
\author[add2]{Karan Singla\corref{cor2}}
\cortext[cor2]{Work done while Karan Singla was at University of Southern California.
}
\author[add3]{Torrey A. Creed}
\author[add4]{David C. Atkins}
\author[add1]{Shrikanth Narayanan}

\address[add1]{Signal Analysis and Interpretation Lab, University of Southern California, Los Angeles, CA, USA}
\address[add2]{Interactions LLC, Los Angeles, CA, USA}
\address[add3]{Department of Psychiatry, University of Pennsylvania, Philadelphia, PA, USA}
\address[add4]{Department of Psychiatry and Behavioral Sciences, University of Washington, Seattle, WA, USA}


\begin{abstract}
Text-based computational approaches for assessing the quality of psychotherapy are being developed to support quality assurance and clinical training. 
However, due to the long durations of typical conversation based therapy sessions, and due to limited annotated modeling resources, computational methods largely rely on frequency-based lexical features or dialogue acts to assess the overall session level characteristics. In this work, we propose a hierarchical framework to automatically evaluate the quality of transcribed Cognitive Behavioral Therapy (CBT) interactions. Given the richly dynamic nature of the spoken dialog within a talk therapy session, to evaluate the overall session level quality, we propose to consider modeling it as a function of local variations across the interaction. To implement that empirically, we  
divide each psychotherapy session into conversation segments and initialize the segment-level qualities with the session-level scores. First, we produce segment embeddings by fine-tuning a BERT-based model, and predict segment-level (local) quality scores.  These embeddings are used as the lower-level input to a Bidirectional LSTM-based neural network to predict the session-level (global) quality estimates. 
In particular, we model the global quality as a linear function of the local quality scores, which allows us to update the segment-level quality estimates based on the session-level quality prediction.
These newly estimated segment-level scores benefit the BERT fine-tuning process, which in turn results in better segment embeddings.
We evaluate the proposed framework on automatically derived transcriptions  from real-world CBT clinical recordings to predict session-level behavior codes.  The results indicate that our approach leads to improved evaluation accuracy for most codes when used for both regression and classification tasks.
\end{abstract}



\end{frontmatter}


\section{Introduction}
\label{sec:into}

Psychotherapy, as a type of mental health treatment, has been developed to assist people with mental illness and overcome problems. The primary method to evaluate a psychotherapy session --- which involves conversations between a therapist and a patient --- is through a process called \emph{behavioral coding} in which trained human raters identify and annotate clinically-relevant global (session-level) or local (utterance-level) behaviors of the participants \citep{bakeman2012behavioral}. However, such a process demands human raters to listen to long audio recordings or read through manually transcribed 
sessions, which leads to a prohibitively high cost in terms of both time and human resources, and hinders therapy evaluation from being widely applied in real-life scenarios \citep{fairburn2011therapist}. 

To overcome such problems related to manual annotation and coding, computational approaches for modeling and assessing the quality of conversation-based behavioral signals \citep{narayanan2013behavioral} have been recently developed and used in multiple clinical domains such as for autism diagnosis \citep{bone2016use}, understanding oncology communication \citep{alam2018annotating, chen2020automated}, and supporting primary care \citep{park2019detecting}. A great amount of work has been particularly focused on psychotherapy interactions, including addiction counseling \citep{can2015dialog, xiao2016behavioral,gibson2017attention, singla2018using, tavabi2020multimodal} and couple therapy sessions \citep{black2013toward, tseng2017approaching}.
Those methods have focused on predicting both utterance-level and globally-coded, session-level behaviors. 
There are different forms of psychotherapy based on the adopted theoretical inclination and scientific foundation, and in how they are practically realized. Among those, 
Cognitive Behavioral Therapy (CBT) is a widely used evidence-based psychotherapy that aims at enabling shifts in the patient's thinking and behavioral patterns \citep{judith2020cognitive}. During a session, the therapist helps the patient to identify inaccurate or unhelpful thoughts and beliefs, evaluate the evidence for those beliefs, and develop new patterns of
thinking that lead to behavioral and emotional change.

The focus of this paper is on the computational assessment of CBT talk sessions.  In CBT the quality of psychotherapy is coded only based on (global) session-level attributes, a task which is especially challenging to be performed automatically because of the inherent complexity of a session, notably due to the long session duration, structured to address multiple varied facets of the therapy process, and the rich diversity across specific client-therapist sessions. The automation is additionally challenged due to limited data resources, especially characterizing specific aspects within a therapy that contribute to the overall judged quality. Previous works on automatically evaluating the CBT session quality hence have adopted coarse features based either on the word frequency \citep{flemotomos2018language} or on the distributions of local behavioral acts \citep{chen2020feature}, which are limited in capturing potentially useful contextual information available within an interaction.   

Automated behavioral coding has benefited from the development of neural network language models. Recently proposed models are able to capture rich contextual information making them widely useful across a variety of domains. Notably, BERT (Bidirectional Encoder Representations from Transformers) models have demonstrated significant improvements in multiple natural language processing (NLP) tasks \citep{devlin2018bert}. However, one of the key challenges related to BERT is its limited capability to handle long sequences due to memory-related limitations. In particular, traditional BERT models can handle sequences of at most 512 tokens, which is 
a serious limitation when processing long, multi-turn conversations, such as exemplified in CBT sessions. To address this problem, researchers have extended the transformers to architectures that can better process long text. Many of these works are based on 
left-to-right autoregressive models which makes them unsuitable for tasks requiring bidirectional information \citep{sukhbaatar2019adaptive, rae2019compressive, dai2020transformer}. Recently, autoencoding models such as \emph{Longformer} \citep{beltagy2020longformer} and \emph{BigBird} \citep{zaheer2020big} were proposed, which combine windowed local-context self-attention and global attention, and can process up to 4,096 input tokens. However, such approaches aggravate the data-hungry problem of adapting the language model to domains with limited resources, while the maximum allowed length is still not sufficient to process real-world clinical sessions. A recent attempt for predicting the CBT codes using deep neural networks in \citep{flemotomos2021automated} employed a recurrent neural network over BERT-based utterance embeddings. However, in this work, BERT was regarded as a feature extractor without task-specific fine-tuning, which might lead to a sub-optimal performance \citep{pappagari2019hierarchical}.

\cite{pappagari2019hierarchical} proposed a transformer-based hierarchical framework for the task of long document classification. The method chunks input documents into blocks, fine-tunes BERT with those blocks to obtain their representations, and then employs a recurrent neural network \citep{rumelhart1986learning} to perform classification. In the present study, we adopt this method as the base configuration to assess CBT quality. However, this approach ignores inherent within-session variability, since, during fine-tuning, all the segments inherit the label of the document that they belong to. To address this limitation, we augment the hierarchical  framework by incorporating an estimator for obtaining accurate {\em local} therapy quality estimates. The overall session quality is then modelled as a linear function of the (local) performance over segments, which allows us to update the segment-level quality estimates based on the session-level (global) quality prediction. More specifically, we approximate the session quality as the weighted average of the local quality estimates and we explore two approaches for determining the weights: 1) estimating them as being proportional to the number of utterances within the segments; 2) learning them through an attention mechanism. We evaluate our framework using real-world CBT session recordings for different session-level behavioral codes and found consistent improvements. Our contributions include:

\begin{itemize}
\item  proposing a novel hierarchical framework incorporating the local quality estimates to model fluctuations of a therapist's performance within a session;
\item modeling the session quality as the weighted average of the local quality and exploring two approaches to determine the segment weights;
\item adapting the transformer-based language model to the psychotherapy domain for both short and long segments;
\item performing both classification and regression tasks for CBT evaluation; this is the first effort for automated CBT evaluation which not only predicts whether a session `is good' but also `how good it is'.
\end{itemize}

\section{Data Description}
\label{sec:data}

The CBT data set used in this work comes from the Beck Community Initiative \citep{creed2016implementation}, a public-academic partnership, and consists of 1,118 sessions with durations ranging from 10 to 90 minutes. Out of those, 292 sessions are accompanied by professional, manual transcriptions, while all the sessions were automatically transcribed by an automated speech pipeline developed for psychotherapy interactions \citep{flemotomos2021ia},  consisting of voice activity detection, diarization, automatic speech recognition, speaker role assignment (therapist vs. patient), and utterance segmentation, which converted speech to punctuated text. We adapted this pipeline to the CBT domain using 100 manually transcribed sessions and evaluated the performance on the remaining 192 ones, with the estimated word error rate being 45.81\%. The error analysis revealed that errors were highly influenced by the presence of speech fillers (e.g., `um', `huh', etc.) and other idiosyncrasies of conversational speech. It should be noted that error rates around this value have been reported to be typical in conversational medical interactions \citep{kodish2018systematic} and inevitably degrade the downstream tasks. From a practical viewpoint, it is important to study the feasibility of applying NLP techniques under real-world circumstances where perfect transcriptions are not available. Several studies have successfully performed NLP tasks on decoded transcripts with similar word error rates across different applications \citep{morchid2016impact, zheng2017navigation}. In our own previous work, we have shown that for the specific problem in the same data domain, the performance degradation of such end-to-end systems due to the automatically derived transcriptions is relatively small when we adopt the frequency-based features~\citep{chen2020feature}.

\begin{figure}[]
\centering
\begin{minipage}[b]{0.15\linewidth}
  \centering
  \centerline{\includegraphics[width=2.4cm]{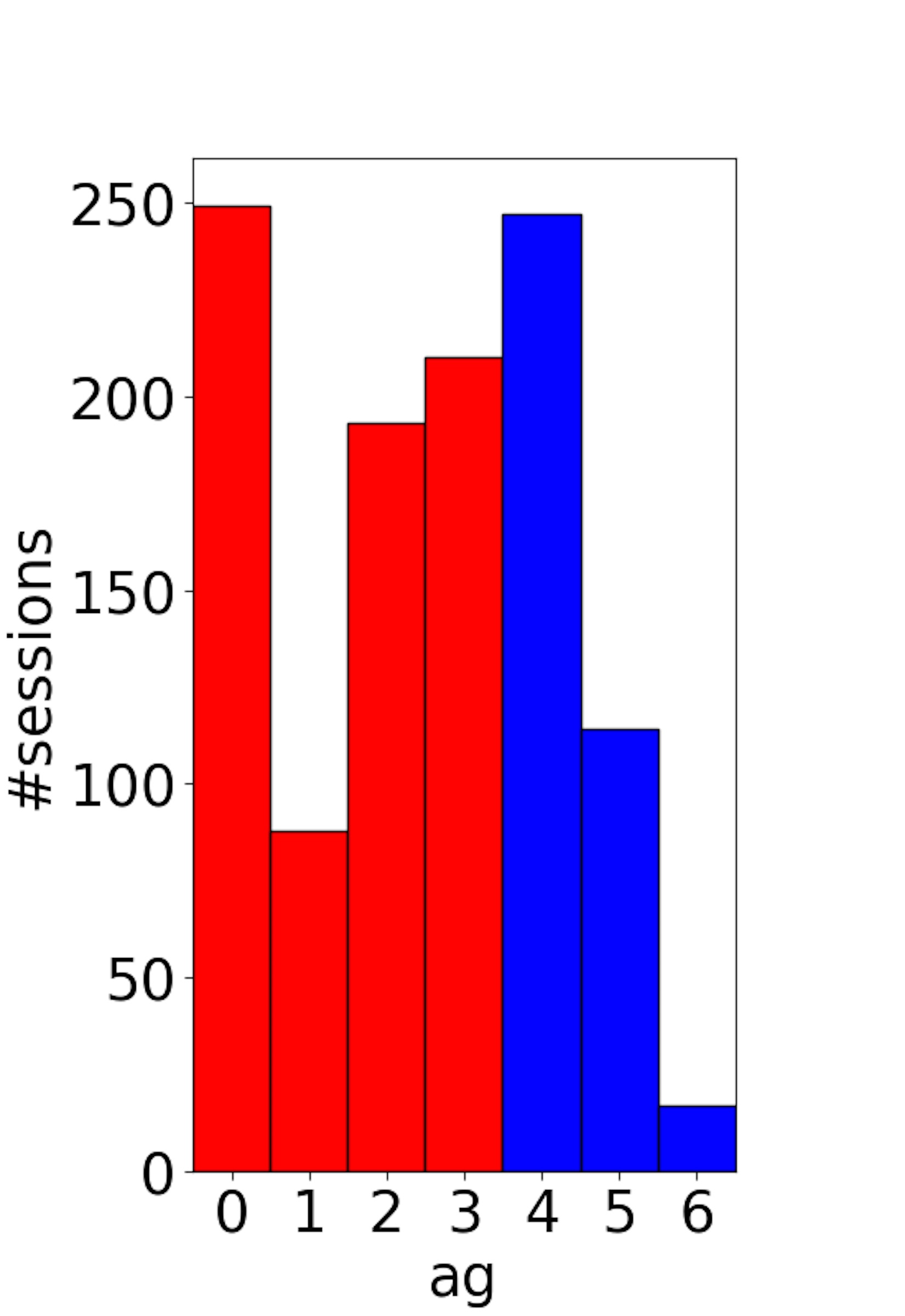}}
\end{minipage}
\begin{minipage}[b]{0.15\linewidth}
  \centering
  \centerline{\includegraphics[width=2.4cm]{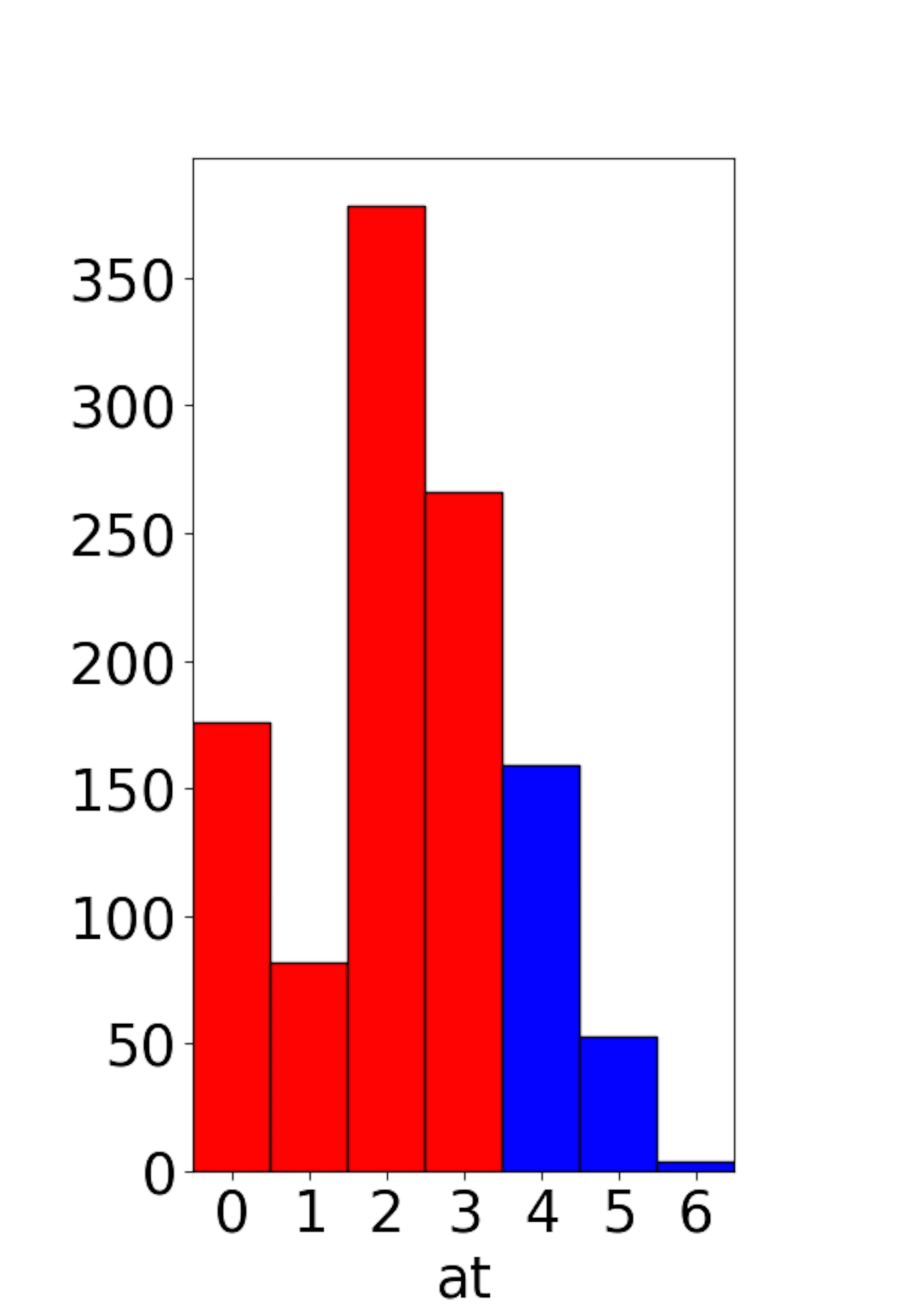}}
\end{minipage}
\begin{minipage}[b]{0.15\linewidth}
  \centering
  \centerline{\includegraphics[width=2.4cm]{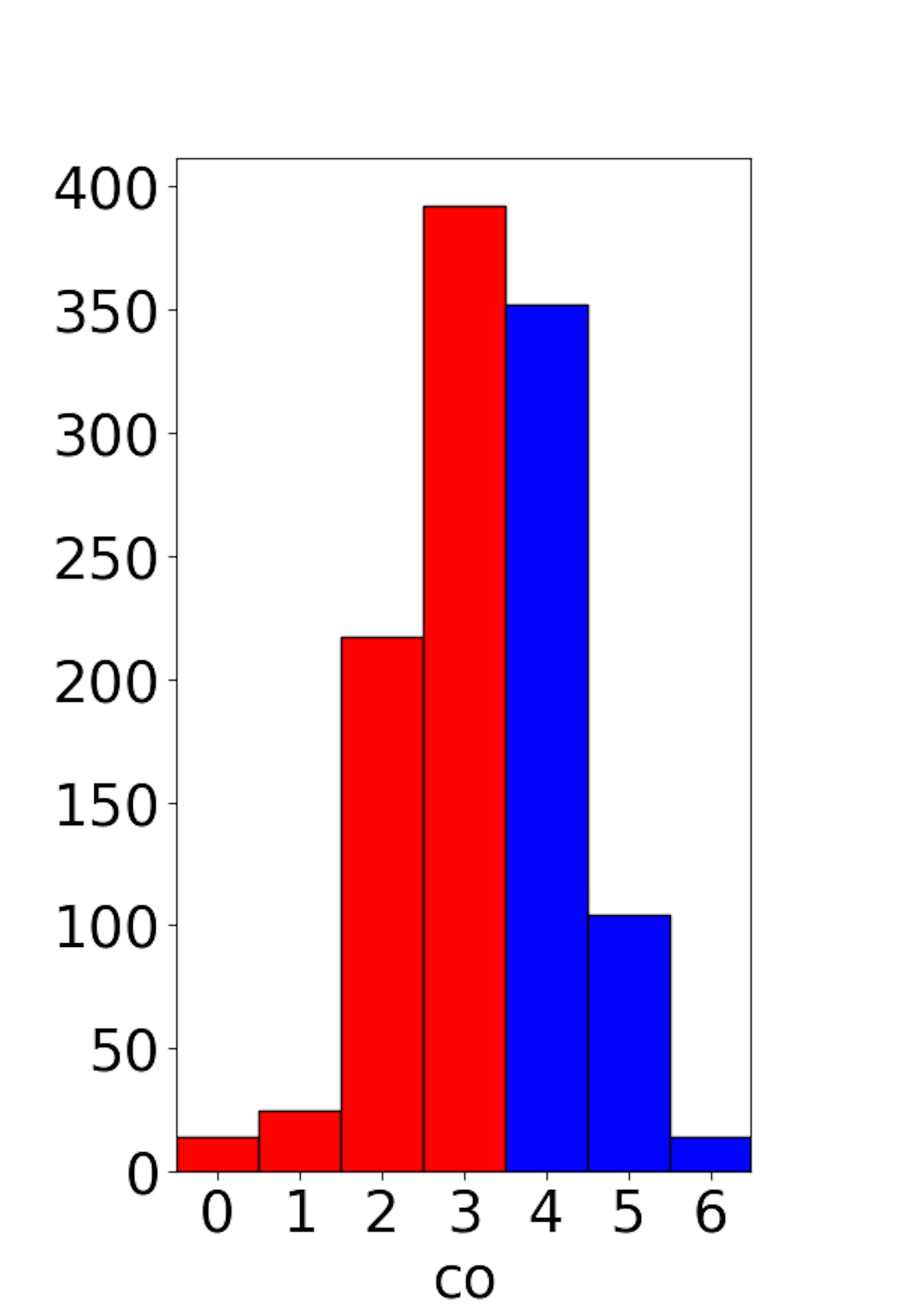}}
\end{minipage}
\begin{minipage}[b]{0.15\linewidth}
  \centering
  \centerline{\includegraphics[width=2.4cm]{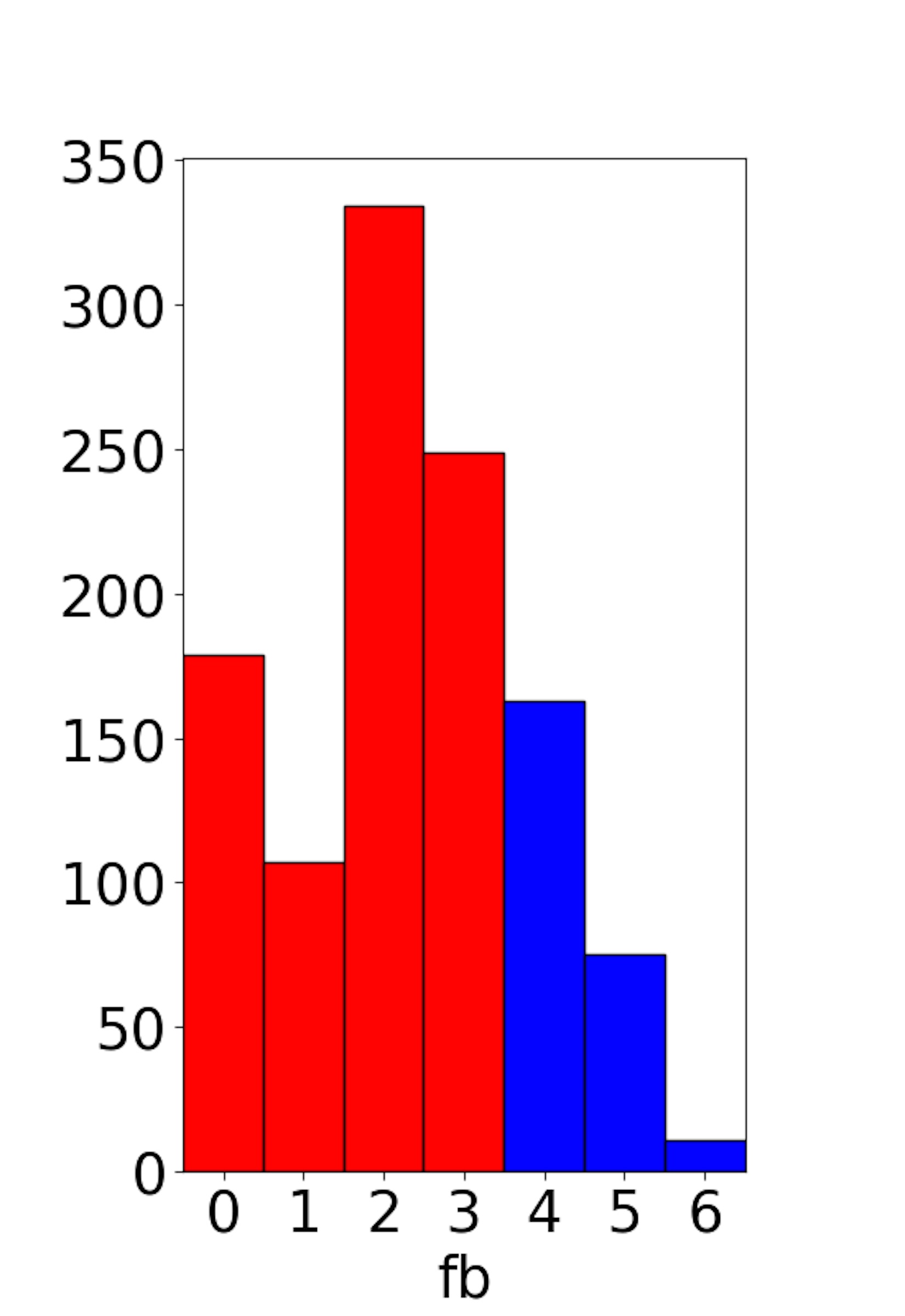}}
\end{minipage}
\begin{minipage}[b]{0.15\linewidth}
  \centering
  \centerline{\includegraphics[width=2.4cm]{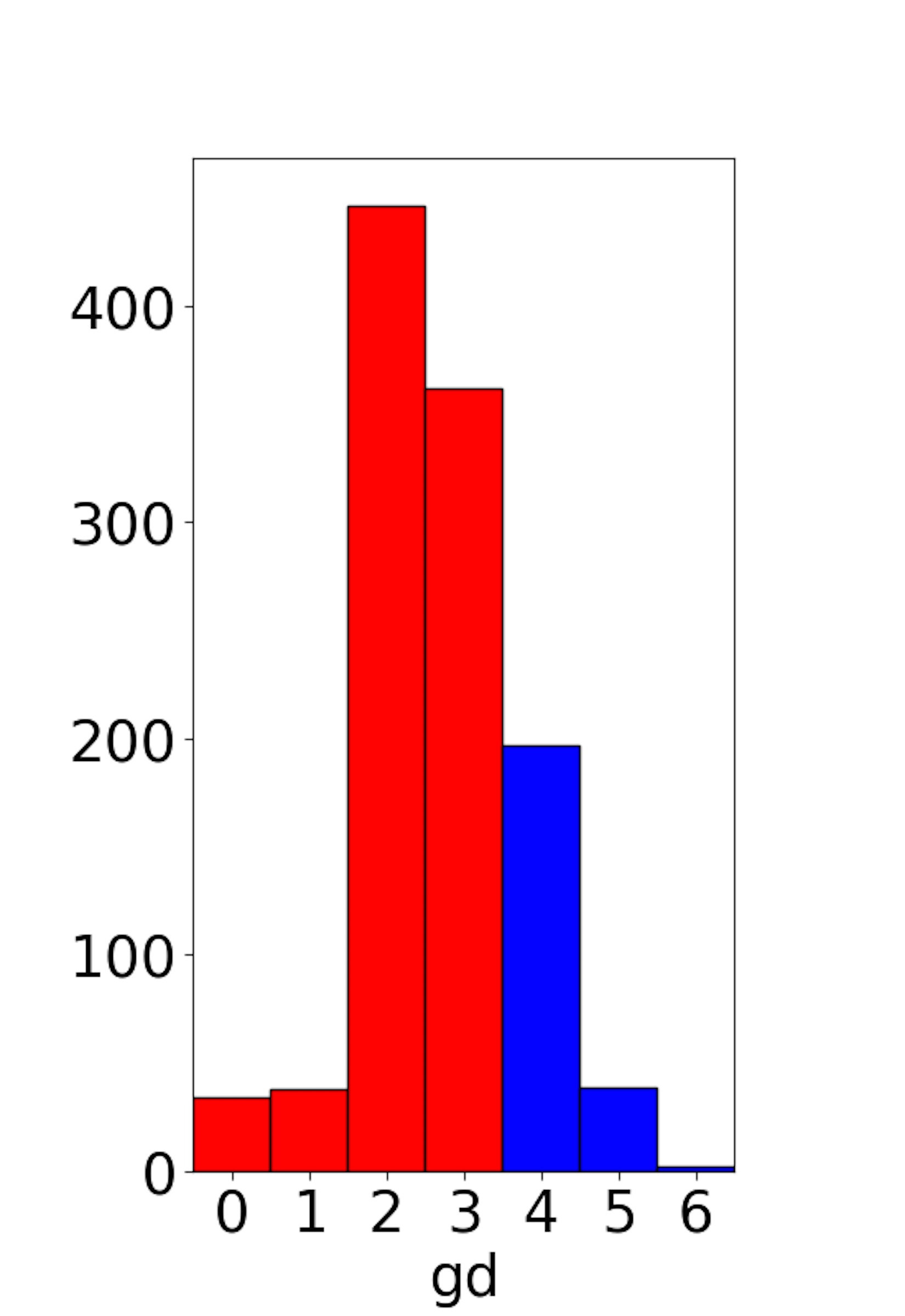}}
\end{minipage}
\begin{minipage}[b]{0.15\linewidth}
  \centering
  \centerline{\includegraphics[width=2.4cm]{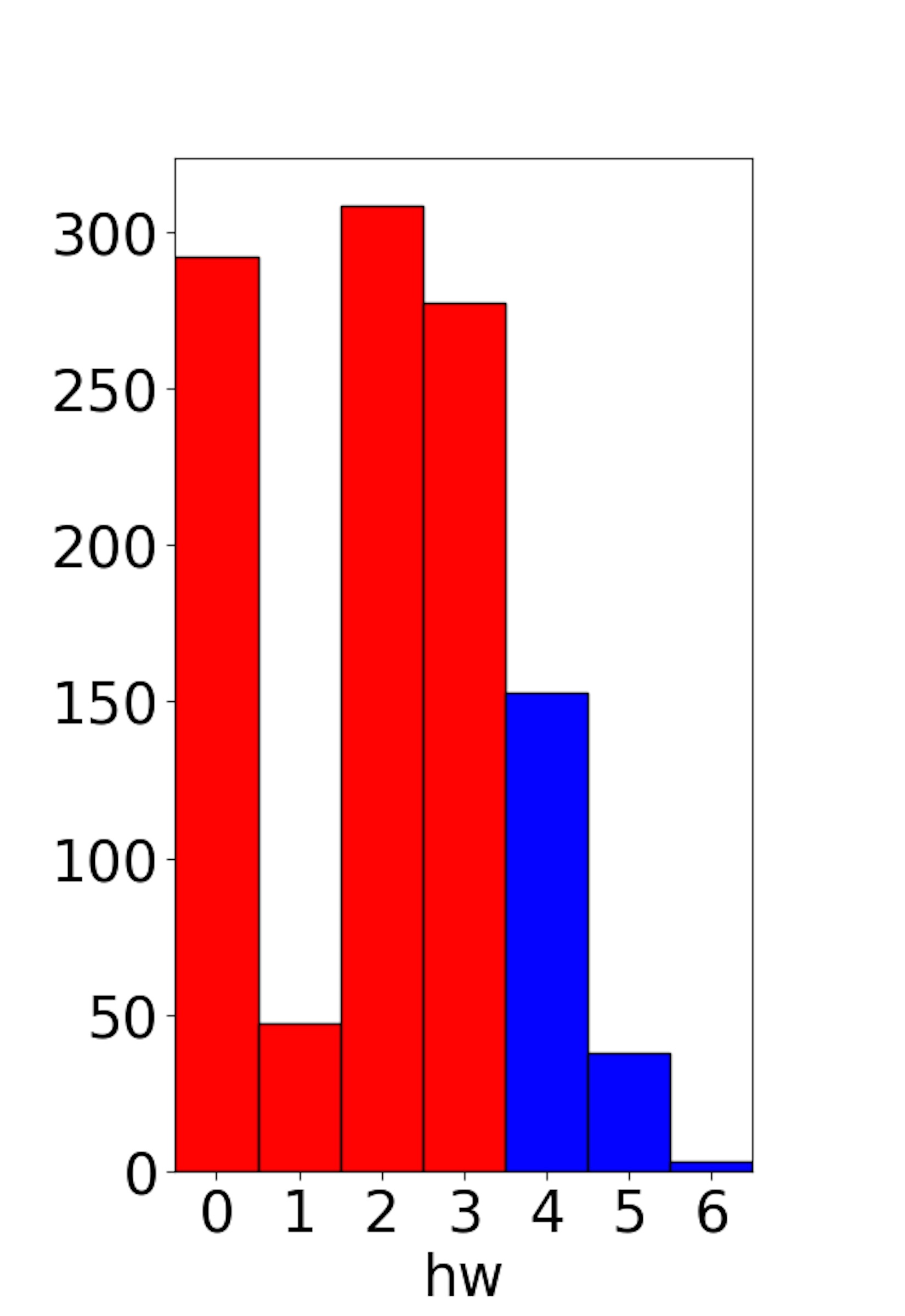}}
\end{minipage}
\begin{minipage}[b]{0.15\linewidth}
  \centering
  \centerline{\includegraphics[width=2.4cm]{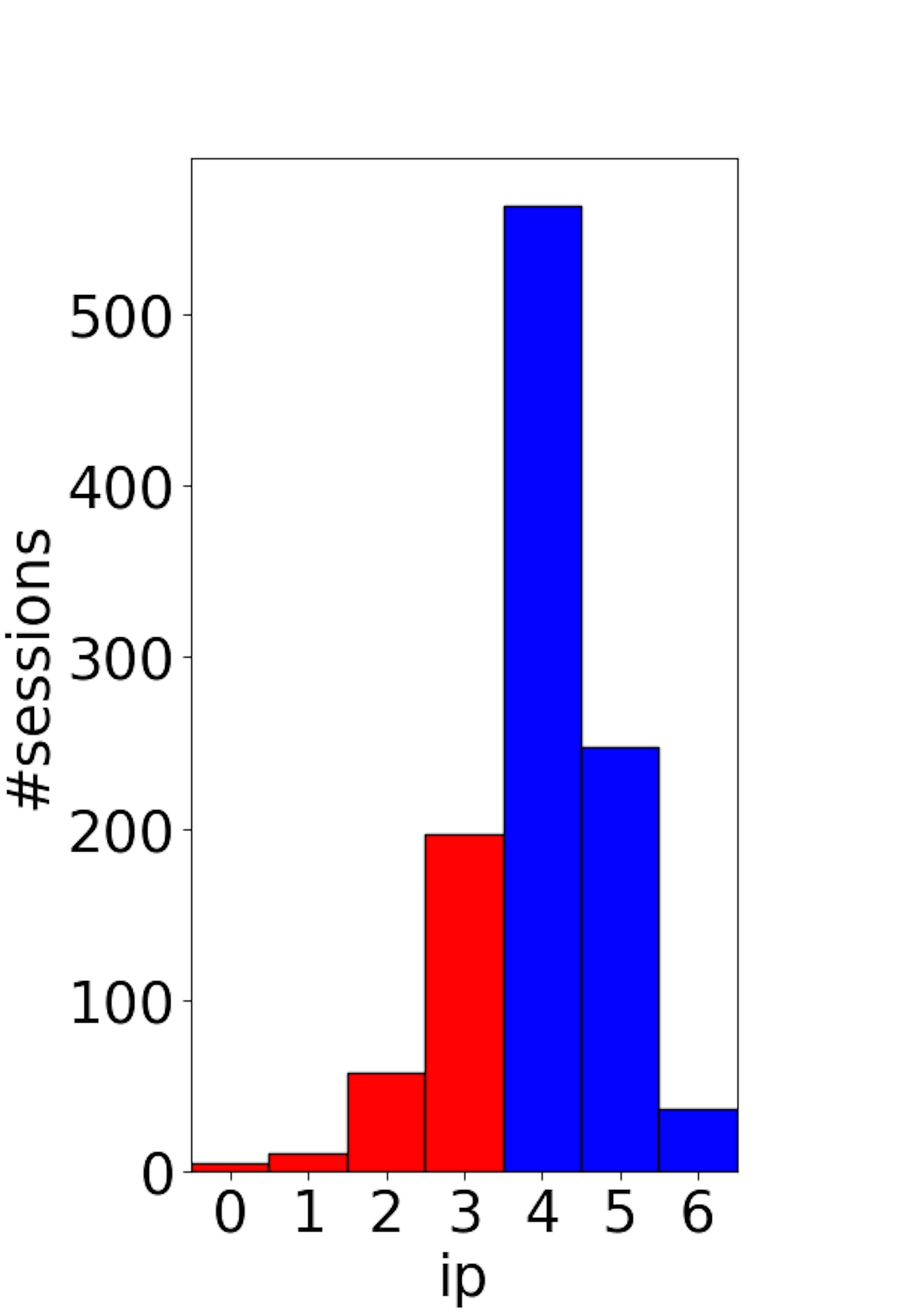}}
\end{minipage}
\begin{minipage}[b]{0.15\linewidth}
  \centering
  \centerline{\includegraphics[width=2.4cm]{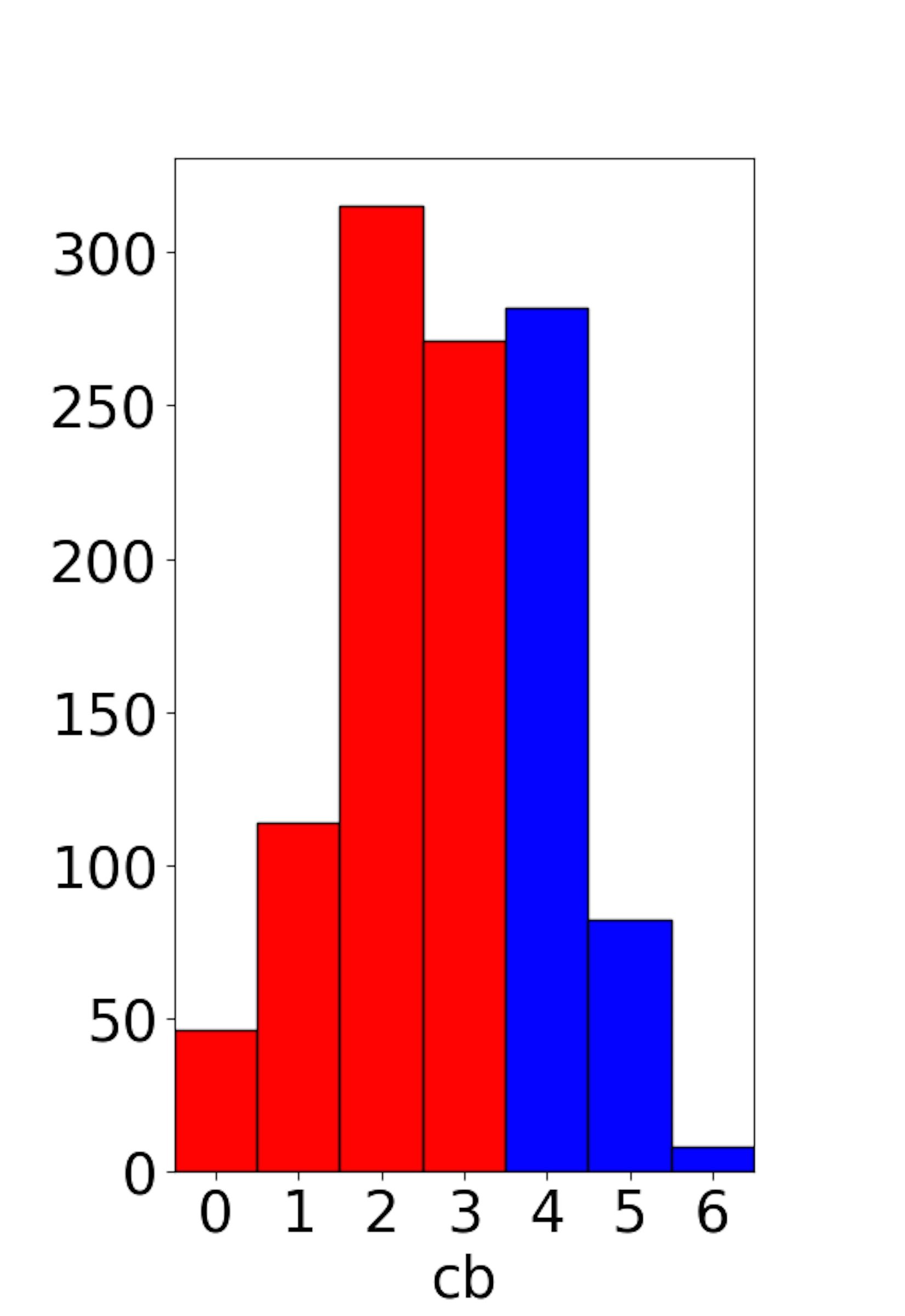}}
\end{minipage}
\begin{minipage}[b]{0.15\linewidth}
  \centering
  \centerline{\includegraphics[width=2.4cm]{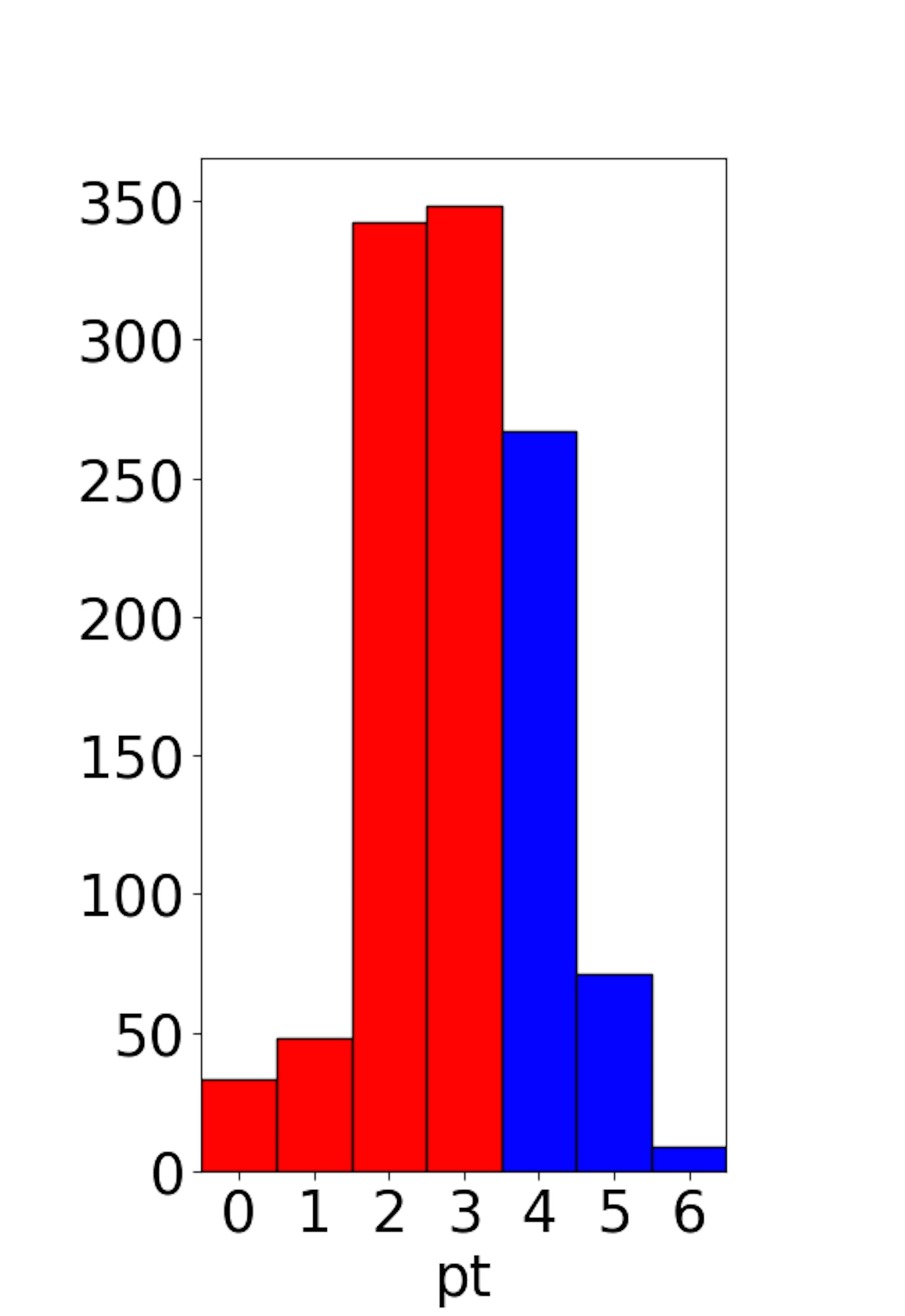}}
\end{minipage}
\begin{minipage}[b]{0.15\linewidth}
  \centering
  \centerline{\includegraphics[width=2.4cm]{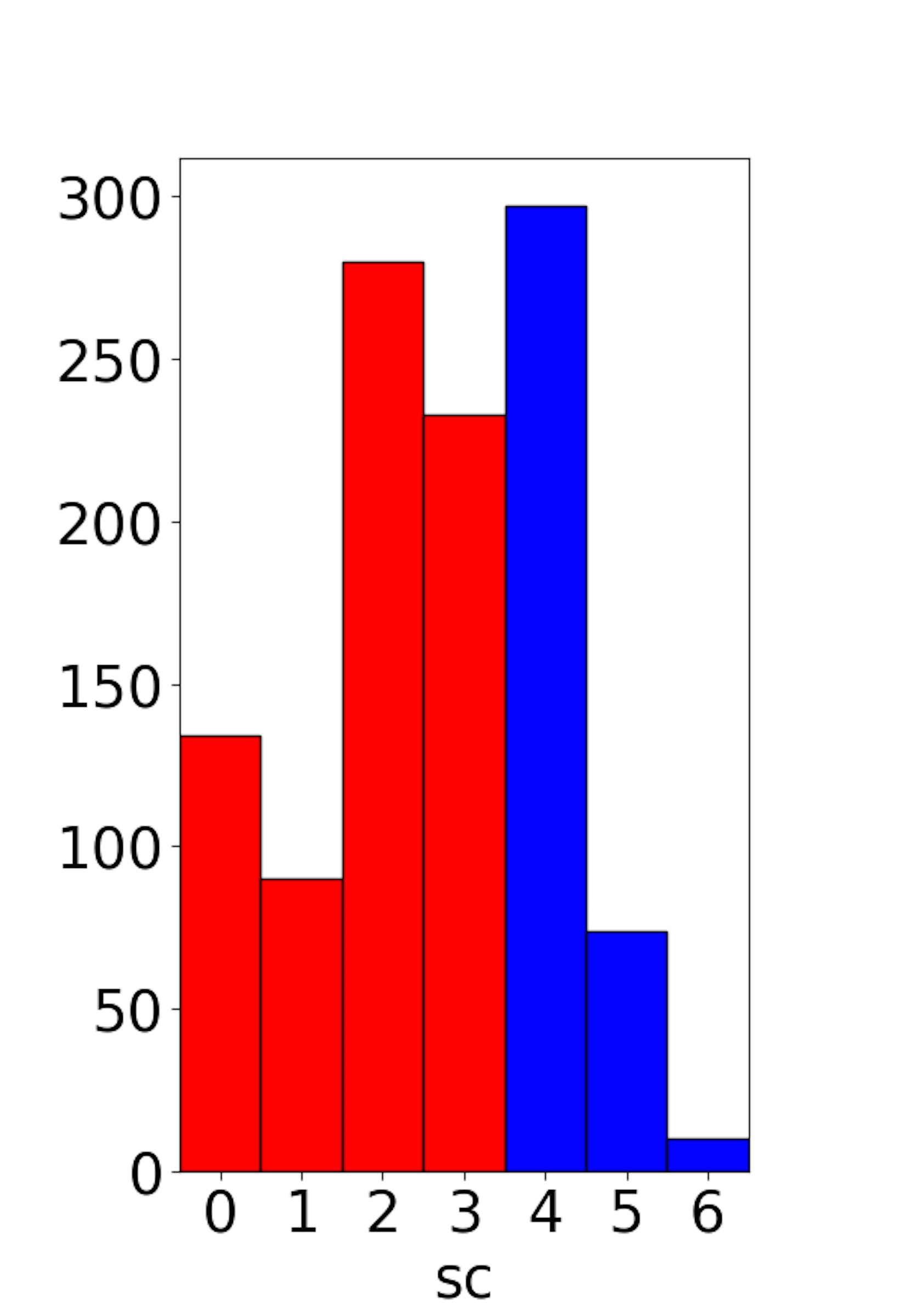}}
\end{minipage}
\begin{minipage}[b]{0.15\linewidth}
  \centering
  \centerline{\includegraphics[width=2.4cm]{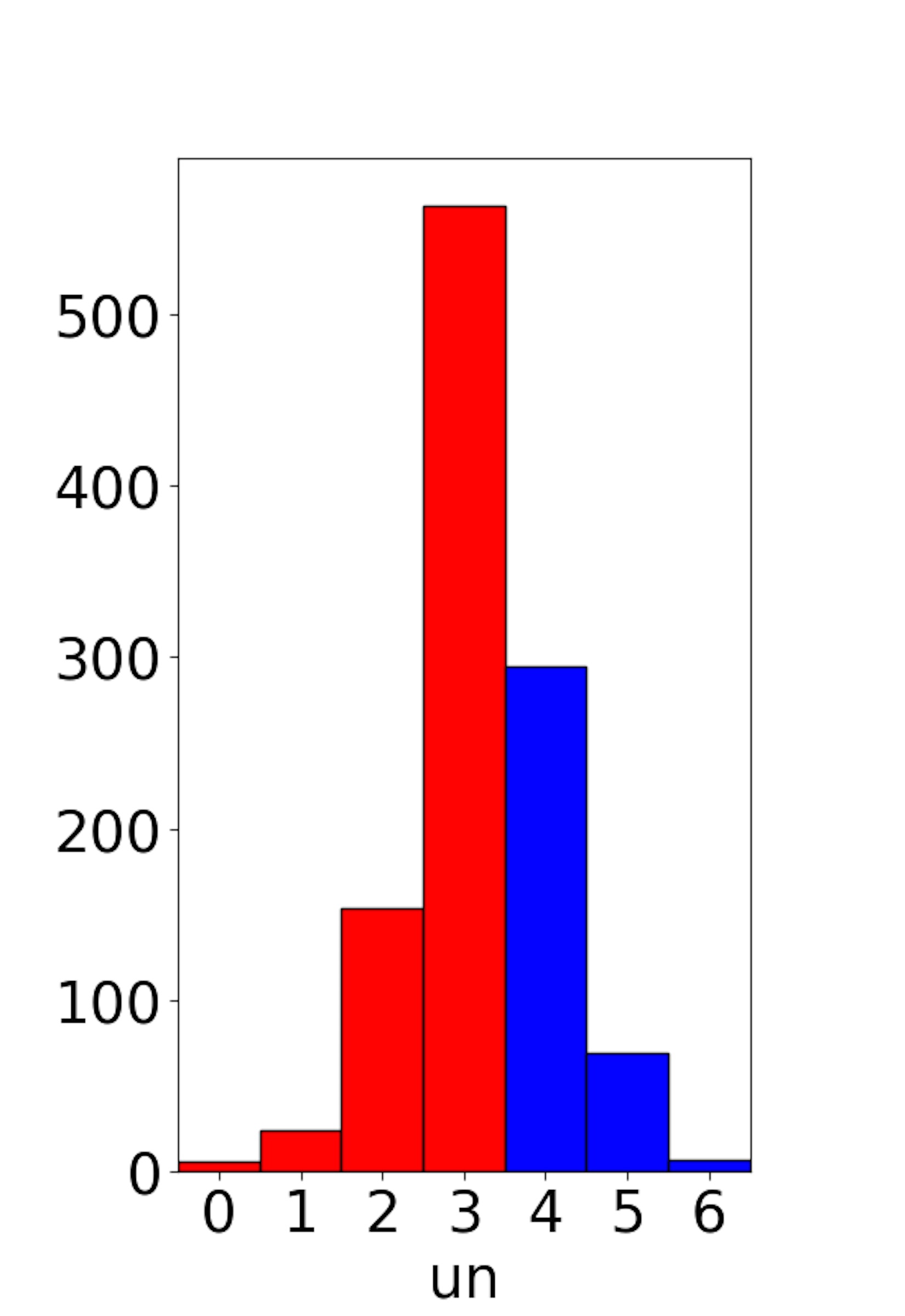}}
\end{minipage}
\begin{minipage}[b]{0.15\linewidth}
  \centering
  \centerline{\includegraphics[width=2.4cm]{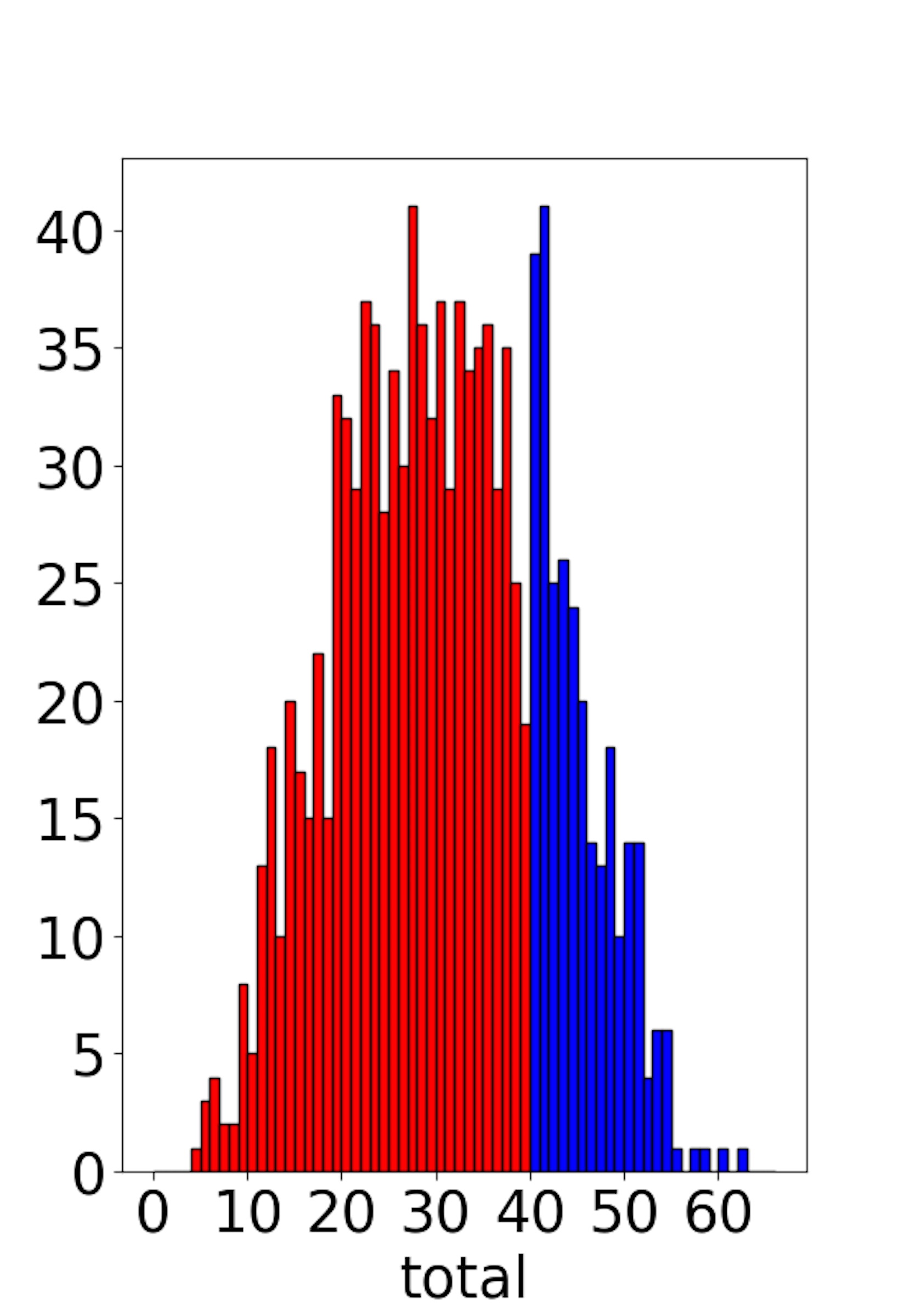}}
\end{minipage}
\caption{Distribution of the 11 CTRS codes (and the total CTRS)}
\label{fig:histogram}
\end{figure}

The entire available data set has been manually annotated to assess session quality. The quality evaluation of CBT is based on the Cognitive Therapy Rating Scale (CTRS, \cite{young1980cognitive}), which defines the session-level behavioral codes shown in Table~\ref{tab:labels}. Each session is evaluated according to 11 codes scored on a 7 point Likert scale ranging from 0 (poor) to 6 (excellent). The sum over all the 11 codes, called the total CTRS score, is typically used as the overall measure of session quality and ranges from 0 to 66. We binarized each CTRS code by assigning codes greater than or equal to 4 as `high' and less than 4 as `low', since 4 is the primary anchor indicating the skill is fully present, but still with room for improvement \citep{young1980cognitive}. For the overall CTRS quality, we binarized it by setting to `high' sessions with total CTRS score greater than or equal to 40 and `low' those with total CTRS less than 40, since a score of 40 is regarded as the benchmark for CBT competence \citep{shaw1999therapist}. The score distributions for different CTRS codes across all the 1,118 sessions are given in Fig.~\ref{fig:histogram}. In this study, a total of 28 doctoral-level CBT experts served as raters. In order to prevent rater drift, they were required to demonstrate calibration before coding, which resulted in high inter-rater reliability (ICC = 0.84, \cite{creed2016implementation}).

\begin{table}[htb]
\caption{CBT behavior codes defined by the CTRS manual}
  \label{tab:labels}
  \centering
\resizebox{0.65\textwidth}{!}{\begin{tabular}{{llll}}
\toprule
\multirow{2}{*}{Abbr.} & \multirow{2}{*}{CTRS code}                     & \multicolumn{2}{l}{`low'/`high' (count) } \\ \cline{3-4}
                       &                                                & Train              & Test              \\ 
                       \midrule
ag                     & agenda                                         & 594/308            & 146/70            \\ 
at                     & application of cognitive behavioral techniques & 735/167            & 167/49            \\ 
co                     & collaboration                                  & 528/374            & 120/96            \\ 
fb                     & feedback                                       & 706/196            & 163/53            \\ 
gd                     & guided discovery                               & 716/186            & 164/52            \\ 
hw                     & homework                                       & 757/145            & 167/49            \\ 
ip                     & interpersonal effectiveness                    & 215/687            & 56/160            \\ 
cb                     & focusing on key cognitions and behaviors       & 615/287            & 131/85            \\ 
pt                     & pacing and efficient use of time               & 635/267            & 136/80            \\ 
sc                     & strategy for change                            & 606/296            & 131/85            \\ 
un                     & understanding                                  & 603/299            & 144/72            \\ 
\midrule
total                  & total score                                    & 683/188            & 156/60   \\ 
\bottomrule
\end{tabular}}
\end{table}

We split the data into training and testing sets with a roughly 80\%:20\% ratio across therapists, with the sessions in the training set being used both for training our models and for domain adaptation. 
The 
label distributions of the binarized codes for both the training and testing sets are shown in Table~\ref{tab:labels}. Besides the CBT data, we used automatically transcribed psychotherapy sessions from a university counseling center (UCC) decoded by the same automated speech pipeline mentioned above to adapt the BERT models. As shown in Table~\ref{tab:dataset}, 
the UCC data set contains more sessions than the CBT one, while the two sets are similar in terms of domain, session duration, and number of words per utterance, which demonstrates its appropriateness for pre-training and adapting the language model.

\begin{table*}[htb]
\caption{Statistics describing the datasets used for the experiments.}\label{tab:dataset}
\begin{center}
\resizebox{0.9\textwidth}{!}{
\begin{tabular}{|c|c|c|c|c|c|}
\hline
{\textbf{Dataset}} & \begin{tabular}[c]{@{}c@{}}sessions\\ (count)\end{tabular} & \begin{tabular}[c]{@{}c@{}}therapists\\ (count)\end{tabular} & \begin{tabular}[c]{@{}c@{}} session duration [min]\\ (mean$\pm$std)\end{tabular} & \begin{tabular}[c]{@{}c@{}}utterances per session\\ (mean$\pm$std)\end{tabular} & \begin{tabular}[c]{@{}c@{}}words per utterance\\ (mean$\pm$std)\end{tabular} \\ \hline
CBT    & 1118 & 383  & 35.5$\pm$12.8 & 656.3$\pm$270.3 & 8.4$\pm$8.3 \\ \hline
UCC   & 4268 & 59  & 38.9$\pm$10.0  & 665.2$\pm$226.2   & 10.2$\pm$11.2 \\ \hline
\end{tabular}}
\end{center}
\end{table*}

\section{Hierarchical framework}
\label{sec:framework}

We partition a CBT session $C$ into $N$ segments $\{C_1, C_2,...,C_N\}$. In order to avoid splitting the session in-between utterances which comprise complete thought units, we divide the session into segments every $M$ utterances. We denote the session-level score of $C$ as $s$ and the local quality corresponding to the segment $C_i$ as ${s_i}$.

\begin{figure}[htb]
  \centering
  \includegraphics[width=10.5cm]{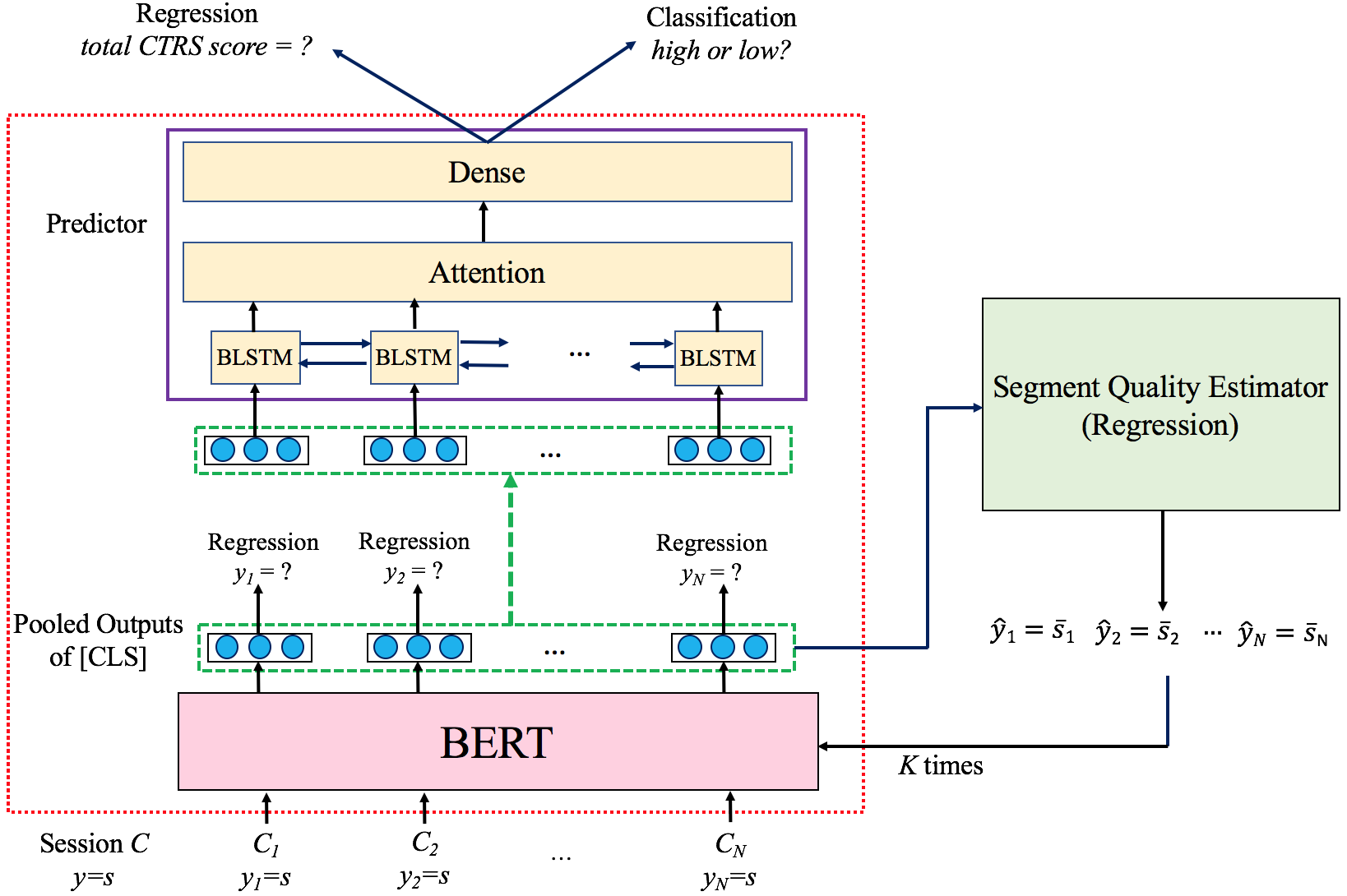}
  \vspace{-0.1cm}
  \caption{Proposed framework using hierarchical transformers.}
  \label{fig:ReHierBERT}
   \vspace{0cm}
\end{figure}

As our base configuration, we adopted the hierarchical transformers structure  presented in the red dashed box of Fig.~\ref{fig:ReHierBERT}, which corresponds to the framework in \citep{pappagari2019hierarchical}. In that case, the segment quality is simply considered to be the score of the session it belongs to. So, every segment is labeled with the score $y_i=s$ and those segment-label pairs are used for BERT fine-tuning. We obtain the segment embeddings by the pooled output (embedding of the initial [CLS] token) from the last transformer block of BERT and feed them into a predictor for session quality evaluation. The predictor includes a bidirectional LSTM layer \citep{hochreiter1997long} to capture dependency features, followed by an additive
self-attention layer \citep{bahdanau20152015}. Different activation functions (linear/sigmoid) of the final dense layer allow for either regression or classification tasks. The loss function for regression is the mean squared error. In the classification scenario, we trained the predictor using the cross-entropy loss with the binary codes (`low' vs. `high' CBT quality) and we assigned weights for each class inversely proportional to their class frequencies.

\begin{algorithm}[htb]
 \caption{Training/Prediction Scheme of Our Hierarchical Framework}\label{alg:alg1}
 \begin{algorithmic}[1]

 \State Initialization: 
 
 Split each conversation $C$ into segments ${C_1, C_2,...C_n}$ 
 
 denote the session-level score of $C$ as $s$ 
 
 initialize the segment quality scores as ${y_1^0=s, y_2^0=s,...,y_N^0=s}$
\State
   \For{$k$ = $0$ to $K$}
   \LongState{Fine-tune BERT by using the local quality scores ${y_1^k, y_2^k,...,y_N^k}$ with a regression task to learn representations of segments.}
   \LongState{Feed segment representations into the LSTM-based model to train the segment quality estimator (SQE) with a regression task.}
   \LongState{Use the trained SQE to predict global quality $\hat{s}$ and local qualities $\hat{s}_1, \hat{s}_2,...,\hat{s}_N$. }
   \State Correcting shift and update local quality: 
   
   $\bar{s}_i= \hat{s}_i + s -\hat{s}$, \; $y_i^{k+1} = \bar{s_i}$
   \EndFor
   \State

   \State Train the predictor with either regression or classification task to make a final prediction of the global quality.
\end{algorithmic}
\end{algorithm}


In this setting, however, all the segments are assigned the label corresponding to the entire session. That means that the behavior of a therapist is assumed to be constant throughout a session,
which is rarely true in the real world. To handle this limitation, we incorporate a local quality estimator which lets us model fluctuations in quality, e.g., of a therapist's performance, within a session.

\section{Framework with local quality estimates}

\subsection{Motivation}

The motivation behind the approach we followed is based on the idea that the score of a session-level behavioral code offers a measure of the overall skill of a therapist over an entire conversation. However, since a psychotherapy session is typically several tens of minutes long, a segment of the conversation by itself often represents a rich and meaningful exchange of ideas and, thus, can be evaluated and scored: the labels (scores) for the segments are interpretable. To that end, we implemented a local quality estimator to estimate the quality of each segment by modeling the global scores as the (weighted) average of local scores.

\subsection{Connecting Session Quality and Segment Quality}



\begin{figure}[htb]
\centering
\begin{minipage}[b]{.49\linewidth}
  \centering
  \centerline{\includegraphics[width=6.5cm]{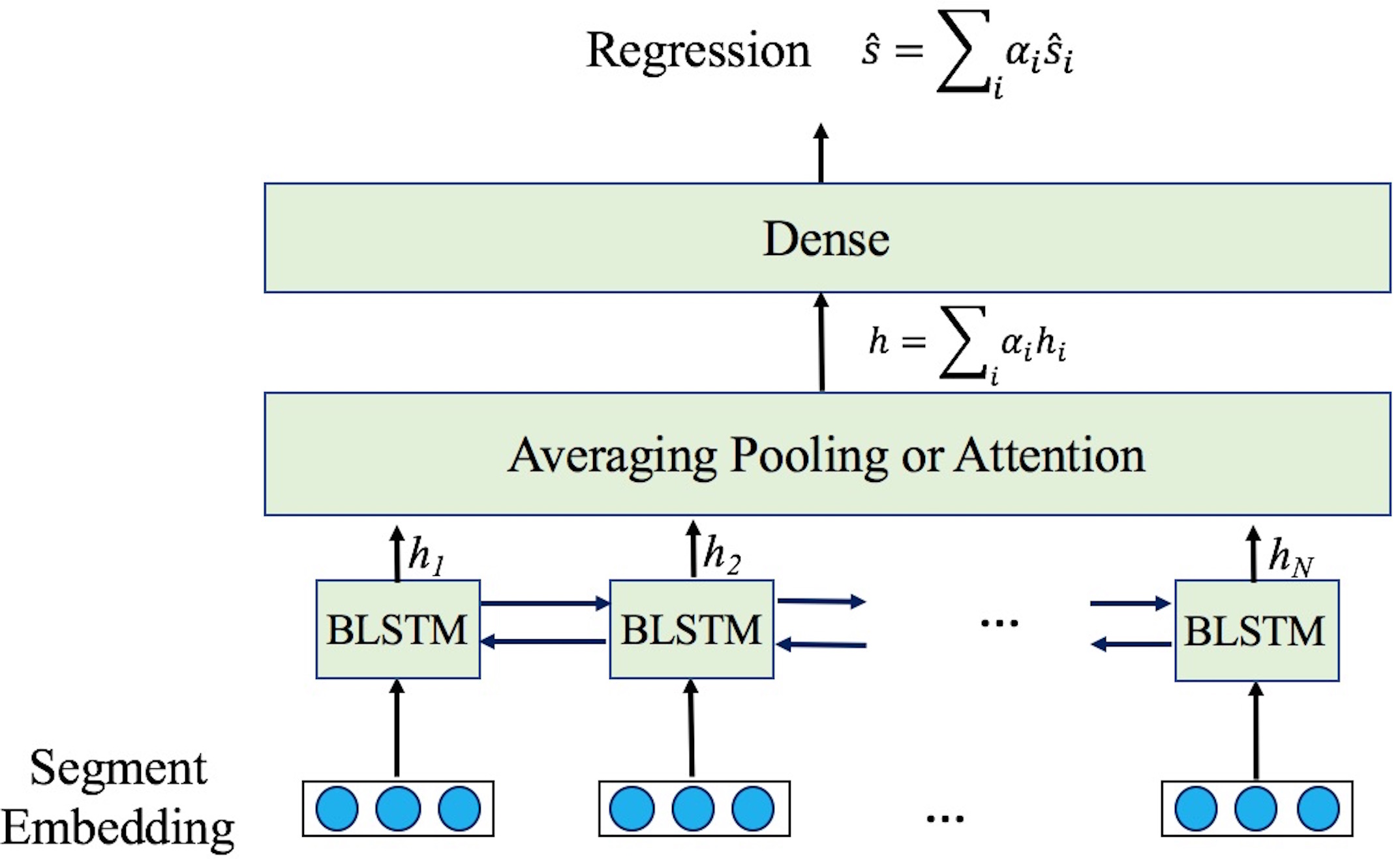}}
  \centerline{\footnotesize{(a) Model \& Training}}\medskip
\end{minipage}
\hfill
\begin{minipage}[b]{0.49\linewidth}
  \centering
  \centerline{\includegraphics[width=6.5cm]{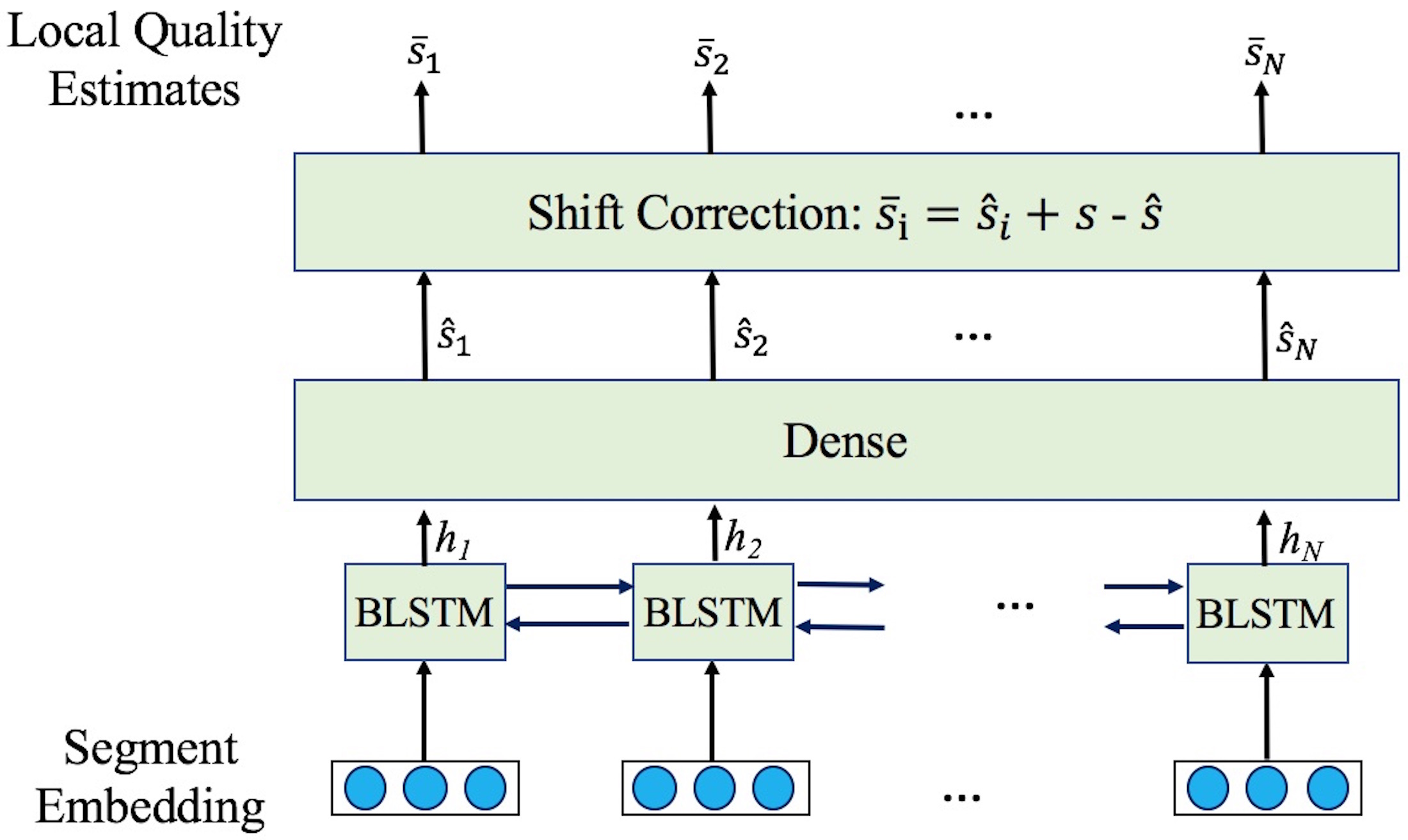}}
  \centerline{\footnotesize{(b) Estimation (fixed model parameters)}}\medskip
\end{minipage}
\caption{Segment Quality Estimator.}
\label{fig:LQE}
\end{figure}

We approximate the session quality $s$ as a weighted average of the performance across the (temporal) segments of an interaction. To estimate the local quality, we implemented the segment quality estimator which is shown in Fig.~\ref{fig:LQE}a. It has a similar structure as the predictor model for a regression task, the only difference being that we replaced the self-attention layer with a linearly activated one
so that its output can be a linear combination of its inputs:

\begin{equation}
h = \sum_i^N \alpha_ih_i \label{eq:2}
\end{equation}
where $h_i$ are the outputs of the BLSTM layers, $h$ is the hidden state fed into the final dense layer, and $\alpha_i$ denote the segment weights.

To determine the segment weights $\alpha_i$, two modes of the segment quality estimator are investigated:

\begin{enumerate}
\item Segment weights are assumed to be proportional to the number of utterances the segments contain, as shown in Eq.~\ref{eq:3}. According to this equation, and based on the way we split each session, the weights of the segment within a session are equal, possibly apart from the last one.\footnote{The last segment might not exactly have $M$ utterances.} We implemented an average pooling layer after the BLSTM (Fig.~\ref{fig:LQE}a). The session quality is approximately equal to the average of its segment qualities. We denote this mode as \emph{even}.

\begin{equation}
\alpha_i=\frac{\text{\#utterances in $C_i$}}{ \text{total \#utterances in $C$}} \label{eq:3}
\end{equation}

\item Segments are assumed to have different importance towards the estimation of the overall session quality, even when they have the same length in terms of the number of utterances. In that case, we implement an attention layer (Fig.~\ref{fig:LQE}a) where the internal context vector learns the segment weights $\alpha_i$ \citep{yang2016hierarchical}, as given by  Eq. \ref{eq:4}. 
The attention layer applies a single-layer MLP to derive the keys $z_i$. Then the attention weights $\alpha_i$ are computed by a softmax function. The internal context vector $u$ is randomly initialized and jointly learned during the training process. The padded sequences are always masked so their attention weights equal to zero. We adopt these attention weights as the segment weights and denote this mode as \emph{uneven} due to the (potentially) uneven distribution of $\alpha_i$.


\begin{equation}\label{eq:4}
\begin{split}
z_i & = tanh(Wh_i+b),\\
\alpha_i & = \frac{exp(z_i^t u)}{\sum_j exp(z_j^t u)},\\
h & = \sum_i^N \alpha_i h_i
\end{split}
\end{equation}

\end{enumerate}

For both modes of the segment quality estimator, the weights of the segments in a session satisfy the condition that $\sum_i^N \alpha_i = 1$.  Having the segment weights determined, the prediction for the session quality is now

\begin{equation}
\hat{s} = \mathcal{L}(h) = \sum_i^N \alpha_i\mathcal{L}(h_i)\triangleq\sum_i^N \alpha_i\hat{s}_i \label{eq:5}
\end{equation}
where $\mathcal{L}$ represents the linear activation of the dense layer. Eq.~\ref{eq:5} indicates we can decompose the global score estimate $\hat{s}$ as the weighted sum of the local score estimates $\hat{s}_i$. As shown in Fig.~\ref{fig:LQE}b, we can obtain these local estimates by feeding the hidden states $h_i$ directly into the dense layer.

In order to account for the deviation between the prediction and the true labels, we update the estimates by correcting the shift (top layer in Fig.~\ref{fig:LQE}b):

\begin{equation}
\bar{s}_i = \hat{s}_i + s - \hat{s}, \;\; i\in{1,...N} \label{eq:6}
\end{equation}
so that

\begin{equation}\label{eq:7}
\begin{split}
\sum_i^N \alpha_i\bar{s}_i &= \sum_i^N \alpha_i(\hat{s}_i + s - \hat{s})
= \sum_i^N \alpha_i\hat{s}_i + s\sum_i^N \alpha_i - \hat{s}\sum_i^N \alpha_i = \hat{s} + s - \hat{s} = s.
\end{split}
\end{equation}
Eq.~\ref{eq:7} indicates that the weighted average of the modified segment quality estimates is equal to the true score of a session.

The complete learning procedure of our hierarchical framework is given in Algorithm \ref{alg:alg1}: we input the updated segment labels $\{y_i=\bar{s_i}\}_{i=1}^N$ in BERT and repeat fine-tuning to get better representations. The loss function for training the segment quality estimator via regression is also the mean squared error. We iterate this process multiple times before employing the predictor to make a final prediction of the overall session quality.

\section{Experiments and Results}

Based on the sequence length distribution on the available dataset, and in order to better exploit the maximum allowed BERT sequence length of 510 tokens, we split each session into sequential segments comprising $M=40$ utterances, with an average sequence length equal to 327.9 words.  
We found that only 3.9\% of the total number of segments in
the CBT dataset we considered were longer than 510 tokens. For comparison, we also 
tried different values for the segment length $M$ and experimentation showed that $M=40$ yields the best results, as explained in Section \ref{subsec:results}.

\subsection{BERT Adaptation}
Due to memory-related limitations (GeForce GTX 1080, 11G) and training efficiency, we adopted a smaller pre-trained uncased BERT variant with 4 layers and 256 hidden states\footnote{https://github.com/google-research/bert} --- denoted as \emph{BERT-small} in this paper --- allowing us to select a larger batch size when training with long sequences.

We adapted \emph{BERT-small} to the CBT domain via domain-adaptive pretraining with the UCC data (Table~\ref{tab:dataset}) and task-adaptive pretraining  \citep{howard2018universal, gururangan2020don} with the training set of CBT over a 90\%-10\% train-eval split. We continued training with the UCC utterances for 1~epoch and CBT utterances for 10 epochs using the following parameters: learning rate of 2e-5, batch size of 64, and sequence length of 64. The adapted model, called \emph{cbtBERT-utt}, achieves an accuracy of 76.6\% on the next sentence prediction task and 44.8\% on the masked language model task.

We continued adapting the model using the long segments ($M=40$) to improve its ability to learn features of longer sequences. Since the number of available segments is very small and some contextual information is filtered, we augmented the segment samples using the following strategy: for each session, we split the transcript 8 times by setting the length of the first segment equal to 5, 10, 15, 20, 25, 30, 35 and 40 utterances. We continued training \emph{cbtBERT-utt} on those segments for 10 epochs with a learning rate of 2e-5, batch size of 8, and sequence length of 512. We denote this adapted model as \emph{cbtBERT-segment}.

\subsection{Experimental Setup}

All our models were implemented in Tensorflow \citep{abadi2016tensorflow} and 20\% of the training data were used for validation. For BERT fine-tuning, we selected the best learning rate (among 1e-5, 2e-5, 3e-5, 4e-5, and 5e-5) on the validation set and used decoupled weight decay regularizer \citep{loshchilov2017decoupled} and a linear learning rate scheduler for optimization. The model was trained for 10 epochs based on the mean squared error loss function with a batch size of 64. The max sequence length was set to 512 tokens for segments of 40 utterances and 64 tokens for segments comprising 1 utterance.\footnote{We also tried setting the max sequence length to 512 tokens for the segments comprising 1 utterance and achieved comparable performance with the results using 64 tokens.}

For the predictor and segment quality estimator models (presented in Fig.~\ref{fig:ReHierBERT} and Fig.~\ref{fig:LQE}), an Adam optimizer \citep{kingma2014adam} was employed with a fixed learning rate of 0.001. The BLSTM layer has the same dimension as the hidden states of BERT. The maximum sequence length was set to 1,600 for short segments and 40 for long segments. We trained the models for up to 50 epochs with a batch size of 64 and an early stopping strategy based on the validation loss. For the regression tasks, we re-scaled the session scores by $f(s)=\left(s-\frac{A}{2}\right)/\frac{A}{2}$ for fast convergence. The value of $A$ equals to 66 for predicting the total CTRS scores and to 6 for predicting each of the CTRS codes so that the normalized labels are always in the range of [-1, 1]. After prediction, 
we map the range [-1, 1] back to [0, 66] for the total CTRS and [0, 6] for the other CTRS codes.

\subsection{Results}\label{subsec:results}

We predict the total CTRS scores via different approaches and report the root mean squared error (RMSE) and mean absolute error (MAE) for the regression tasks and the macro-averaged
F1 score for the classification tasks.

As a baseline, we perform linear regression (LR) and support vector machine (SVM) based classification, coupled with unigrams under a term frequency-inverse document frequency (tf-idf) transformation, which was reported to achieve the best results for the task in \citep{flemotomos2018language}. We denote these two methods by \textbf{tf-idf + LR} and \textbf{tf-idf + SVM}. We also compare the results of our hierarchical framework to the model's performance when replacing the BERT embeddings by the segment-level averaged glove embeddings \citep{pennington2014glove} or by paragraph vectors (doc2vec, \cite{dai2015document}). We extract these segment-level features and directly feed them into the LSTM-based predictor in Fig.~\ref{fig:ReHierBERT}; we denote those approaches by  
\textbf{glove + LSTM} and \textbf{doc2vec + LSTM}. Besides, we perform the evaluation tasks using the pre-trained \emph{BERT-small}, and two sparse attention transformer-based models --- \emph{Longformer} \citep{beltagy2020longformer} and \emph{BigBird} \citep{zaheer2020big} --- by truncating the document to the maximum sequence length.

For our framework, we additionally evaluate the performance with respect to whether fine-tuning BERT with segment scores or not is involved, and with respect to the number of times $k$ we call the segment quality estimator for updating those scores. If $k=0$, the structure is equivalent to the normal hierarchical framework described in Section~\ref{sec:framework} and introduced in \citep{pappagari2019hierarchical}. Furthermore, we report and make comparisons between the two proposed modes of the segment quality estimator: \emph{even} --- in each session, every segment contributes equally towards the session quality --- and \emph{uneven} --- in each session, the contribution of each segment towards the session quality is learnt through an attention mechanism. 
Based on the pre-trained language model we use, the approaches are named as \textbf{BERT-small + LSTM}, \textbf{BERT-cbt-utt + LSTM} and \textbf{BERT-cbt-segment + LSTM}.

The experimental results of the different approaches are shown in Table~\ref{tab:result1}. For all the methods, the difference between RMSE and MAE is relatively small which indicates that outliers do not greatly affect the predictions. Based on the results, we observe that, for evaluating such lengthy conversations, frequency-based methods perform better than simple neural network methods.
The transformer-based model --- \emph{BERT-small} --- achieves a low performance because it only receives 512 tokens for each session, ignoring most of the information. The transformer models with sparse attentions --- \emph{Longformer} and \emph{BigBird} --- increase the maximum possible input sequence length to 4,096 and evaluate the session quality more accurately. However, this length is still not sufficient for CBT conversations, and, as a result, the performance is worse when compared to the hierarchical approaches.
The results of the hierarchical framework suggest that fine-tuning substantially improves the performance when the segment is long. However, fine-tuning is not as effective and might even lead to performance degradation if single utterances are used as segments, since assigning the global session quality to very short chunks of text may result in inaccuracies. Specifically for the \textbf{BERT-small + LSTM} configuration, we experimented with various values $M$ for the segment length and found that $M=40$ resulted to the best results (more details are given in \ref{sec:M_v}). This is why we used this value for the subsequent experiments.
Comparing the various pre-trained BERT models, we conclude that adapting the language model with in-domain data leads to improved prediction performance. Additionally, when using long segments as inputs, \emph{cbtBERT-segment} consistently yields better results than \emph{cbtBERT-utt}, which confirms its suitability for handling longer sequences.

\begin{table}[]

\caption{Evaluation results for total CTRS scores, $M$: \#utterances/segment, $k$: \#times processing the segment scores estimator, SQE: segment quality estimator}
  \label{tab:result1}
  \vspace{0.3cm}
  \centering
\resizebox{0.98\textwidth}{!}{
\begin{tabular}{|l|c|c|c|c|c|c|}
\hline
\multicolumn{1}{|c|}{Approach}           & BERT fine-tune & M  & k & SQE mode & RMSE/MAE           & F1 score (\%) \\ \hline
\multicolumn{7}{|l|}{\textbf{Frequency-based Methods}}                                                             \\ \hline
tf-idf + LR                              & -              & -  & - & -        & 9.48/7.49          & -             \\ \hline
tf-idf + SVM                             & -              & -  & - & -        & -                  & 69.0          \\ \hline
\multicolumn{7}{|l|}{\textbf{Neural Network Methods}}                                                              \\ \hline
\multirow{2}{*}{glove + LSTM}            & -              & 1  & - & -        & 10.05/8.09         & 59.6          \\ \cline{2-7} 
                                         & -              & 40 & - & -        & 9.90/7.99         & 60.2          \\ \hline
\multirow{2}{*}{doc2vec + LSTM}          & -              & 1  & - & -        & 9.88/7.91          & 62.2          \\ \cline{2-7} 
                                         & -              & 40 & - & -        & 9.75/7.80          & 63.0          \\ \hline
\multicolumn{7}{|l|}{\textbf{Transformer Models}}                                                             \\ \hline
BERT-small                              & -              & -  & - & -        & 9.89/7.93          & 61.9             \\ \hline
Longformer                             & -              & -  & - & -        & 9.35/7.31                  & 67.9          \\ \hline
BigBird                             & -              & -  & - & -        & 9.30/7.25                  & 68.5          \\ \hline
\multicolumn{7}{|l|}{\textbf{Hierarchical Framework}$^1$}                                                                       \\ \hline
\multirow{4}{*}{BERT-small + LSTM$^2$ }    & \XSolidBrush              & 1  & 0 & -        & 9.78/7.82          & 62.6          \\ \cline{2-7} 
                                         & \Checkmark              & 1  & 0 & -        & 9.88/7.89          & 62.2          \\ \cline{2-7} 
                                         & \XSolidBrush              & 40 & 0 & -        & 9.68/7.70          & 63.5          \\ \cline{2-7} 
                                         & \Checkmark              & 40 & 0 & -        & 8.78/6.97          & 70.7          \\ \hline
\multirow{4}{*}{BERT-cbt-utt + LSTM}     & \XSolidBrush              & 1  & 0 & -        & 9.57/7.59          & 65.3          \\ \cline{2-7} 
                                         & \Checkmark             & 1  & 0 & -        & 9.56/7.67          & 64.6          \\ \cline{2-7} 
                                         & \XSolidBrush              & 40 & 0 & -        & 9.45/7.50          & 65.5          \\ \cline{2-7} 
                                         & \Checkmark              & 40 & 0 & -        & 8.59/6.80          & 72.0          \\ \hline
\multirow{2}{*}{\begin{tabular}[c]{@{}c@{}}BERT-cbt-segment + LSTM\end{tabular}} & \XSolidBrush              & 40 & 0 & -        & 9.27/7.29          & 67.9          \\ \cline{2-7} 
                                         & \Checkmark              & 40 & 0 & -        & 8.47/6.59          & $\;$ 73.0 $^+$                \\ \hline
\multirow{6}{*}{\begin{tabular}[c]{@{}c@{}}BERT-cbt-segment + LSTM \\ + SQE\end{tabular}} & \Checkmark              & 40 & 1 & Even     & 8.19/6.35          & 74.7          \\ \cline{2-7} 
                                         & \Checkmark              & 40 & 1 & Uneven   & 8.25/6.40          & 74.3          \\ \cline{2-7} 
                                         & \Checkmark              & 40 & 2 & Even     & 8.12/6.29          & 75.0          \\ \cline{2-7} 
                                         & \Checkmark              & 40 & 2 & Uneven   & 8.22/6.38          & 74.5          \\ \cline{2-7} 
                                         & \Checkmark              & 40 & 3 & Even     & \textbf{8.09/6.27} & \textbf{$\;$ 75.1 $^\ast$} \\ \cline{2-7} 
                                         & \Checkmark              & 40 & 3 & Uneven   & 8.22/6.37          & 74.5          \\ \hline
\end{tabular}}
\\
\vspace{0.0cm}
\footnotesize{$^1$Approachs without fine-tuning correspond to the single task models in \cite{flemotomos2021automated}}\\
\footnotesize{$^2$Corresponds to the framework in \cite{pappagari2019hierarchical}}\\
\footnotesize{$*$ is significantly higher than $+$ at $p < 0.05$ based on Student's t-test.}\\
\end{table}

The performance of the hierarchical transformers framework is significantly improved by incorporating a segment quality estimator (SQE), which also outperforms other approaches in Table~\ref{tab:result1}. Moreover, we find that the best results are achieved when we select the ``even" mode for the segment quality estimator, which indicates that, for predicting the total CTRS scores, it is reasonable to model the global quality as the average of the local quality scores.
The results also show that the system performance is improved as we increase the number $k$ of SQE updates, although a plateauing trend is observed.

To further compare the two segment quality estimator modes, we predict each of the CTRS codes with both modes and also present the performance without incorporating a segment quality estimator. The segment length for these experiments is set to 40 utterances ($M=40$) and the results are reported in Table~\ref{tab:result2}.
For both modes, the number of times updating the segment quality estimator is set to one ($k=1$). The `None' column in the table corresponds to performing the tasks without a segment scores estimator ($k=0$). From the table, we can observe that using a segment scores estimator, regardless of the mode selected, leads to improved prediction performance for the majority of the CTRS codes. The segment quality estimator with the ``even" mode yields the best results for 8 out of 11 codes.
However, for the codes \emph{agenda} and \emph{homework}, the ``uneven" mode leads to more accurate predictions. 
We assume that which mode of segment quality estimator achieves better performance depends on inherent characteristics of the codes and the ability of the employed attention mechanism to robustly learn the segment weights. It seems that, for most of the codes, all the segments are almost equally important,
and the uniformly distributed segment weights used by the `even' mode are more accurate than the estimated weights learned by the attention mechanism. For codes like \emph{agenda} and \emph{homework}, the global quality is mainly associated to only a few segments within the session, so the estimated weights using the attention mechanism are more accurate.
We further analyze this behavior in Sec~\ref{sec:att}.

\begin{table}[]
\caption{Comparison of different segment quality estimator modes (for $M=40$), update the segment quality scores for $k=1$ time. None: without using a segment quality estimator}
\vspace{0.3cm}
  \label{tab:result2}
  \centering
  \resizebox{0.8\textwidth}{!}{
\begin{tabular}{|c|c|c|c|c|c|c|}
\hline
\multirow{2}{*}{CTRS code} & \multicolumn{3}{c|}{MAE/RMSE}                       & \multicolumn{3}{c|}{F1 scores (\%)}           \\ \cline{2-7} 
                           & None      & Even               & Uneven             & None          & Even          & Uneven        \\ \hline
ag                         & 0.86/1.10 & 0.85/1.10 & \textbf{0.81/1.05}          & $\;$ 74.6 $^+$          & 74.6 & \textbf{$\;$ 76.6 $^*$}          \\ \hline
at                         & 0.95/1.20 & \textbf{0.90/1.16}          & 0.93/1.18 & $\;$ 64.5 $^+$         & \textbf{ $\;$ 66.3 $^*$}          & 65.6 \\ \hline
co                         & 0.74/0.97 & \textbf{0.73/0.96}          & 0.75/0.98 & 69.6          & \textbf{70.5}          & 68.9 \\ \hline
fb                         & 0.97/1.20 & \textbf{0.89/1.13}          & 0.93/1.15 & $\;$ 66.5 $^+$          & \textbf{$\;$ 69.5 $^*$}          & 68.0 \\ \hline
gd                         & 0.74/0.98 & \textbf{0.71/0.95}          & 0.73/0.98 & $\;$ 63.2 $^+$          & \textbf{$\;$ 66.6 $^*$}          & 64.4 \\ \hline
hw                         & 1.00/1.24 & 0.97/1.21 & \textbf{0.96/1.20}          & 67.0          & 68.2  & \textbf{68.7}         \\ \hline
ip                         & 0.69/0.90& \textbf{0.67/0.89}          & 0.69/0.90          & 60.5 & \textbf{61.5}           & 60.3          \\ \hline
cb                         & \textbf{0.89/1.10} & 0.91/1.12          & 0.91/1.13 & \textbf{69.8}          &  69.3         & 69.3 \\ \hline
pt                         & 0.84/1.16 & \textbf{0.82/1.13}          & 0.84/1.16 & 64.0          & \textbf{65.5}          & 64.4 \\ \hline
sc                         & 1.00/1.25 & \textbf{0.96/1.19}          & 0.98/1.20 & $\;$ 66.9 $^+$          & \textbf{$\;$ 69.4 $^*$}         & 68.6  \\ \hline
un                         & 0.66/0.87 & \textbf{0.62/0.83}         & 0.64/0.85 & 60.2          & \textbf{62.4}          & 61.5 \\ \hline
\end{tabular}}
\\
\footnotesize{$*$ is significantly higher than $+$ at $p < 0.05$ based on Student's t-test.}
\end{table}

\section{Analysis and Discussion of Experimental Results}
\label{sec:dis}

In this section, we investigate the contribution of each segment via the analysis of the attention weights and discuss how the segment quality estimates benefit the prediction of the session-level scores.


\subsection{Attention weights}
\label{sec:att}

\begin{figure}[]
  \centering
  \includegraphics[width=12cm]{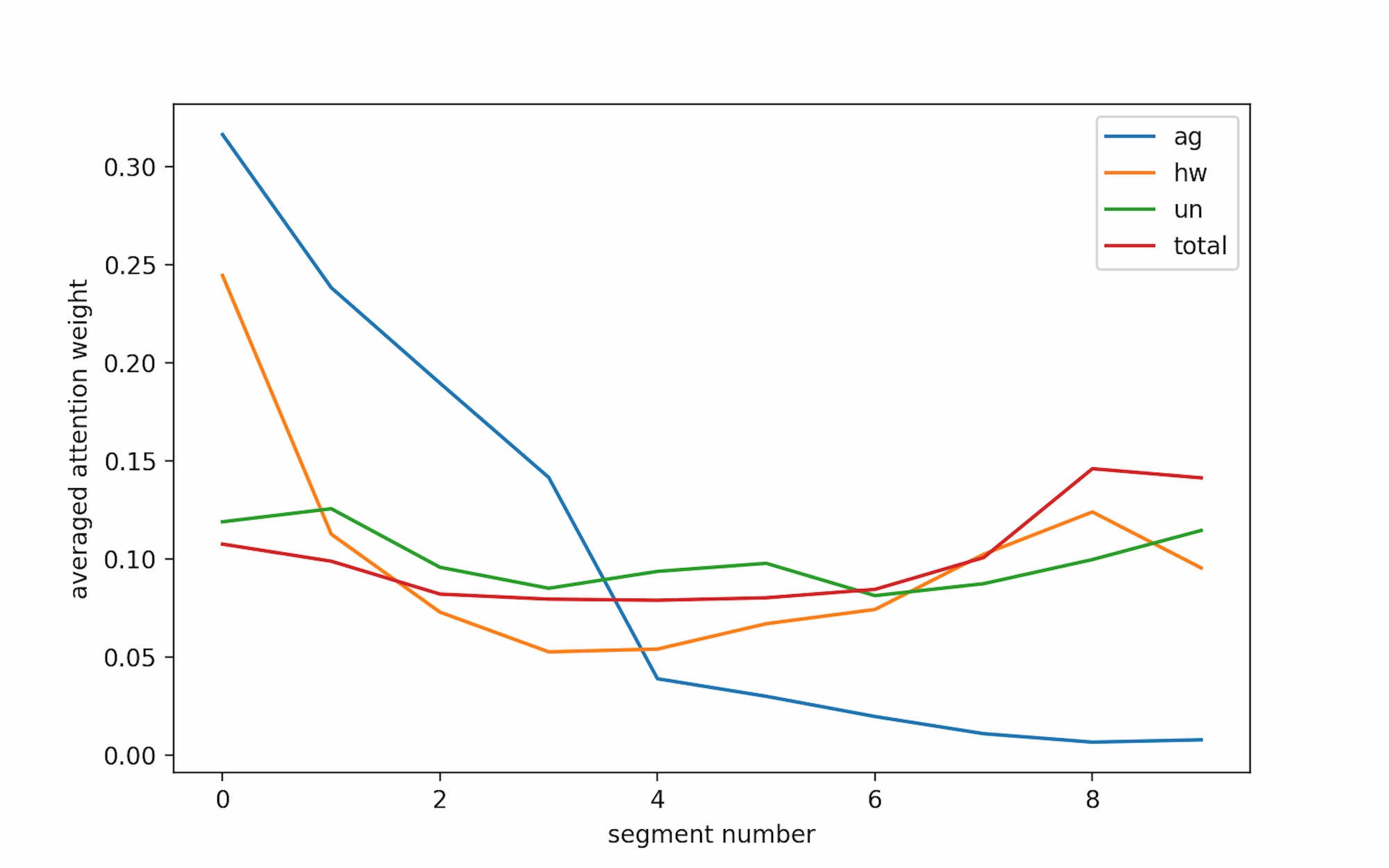}
  \caption{Mean attention weights across the sessions consisting of exactly 10 long segments (of 40 utterances each) in the testing set.}
  \label{fig:attention}
   \vspace{0.0cm}
\end{figure}

\begin{figure}[htb]
  \centering
  \includegraphics[width=10cm]{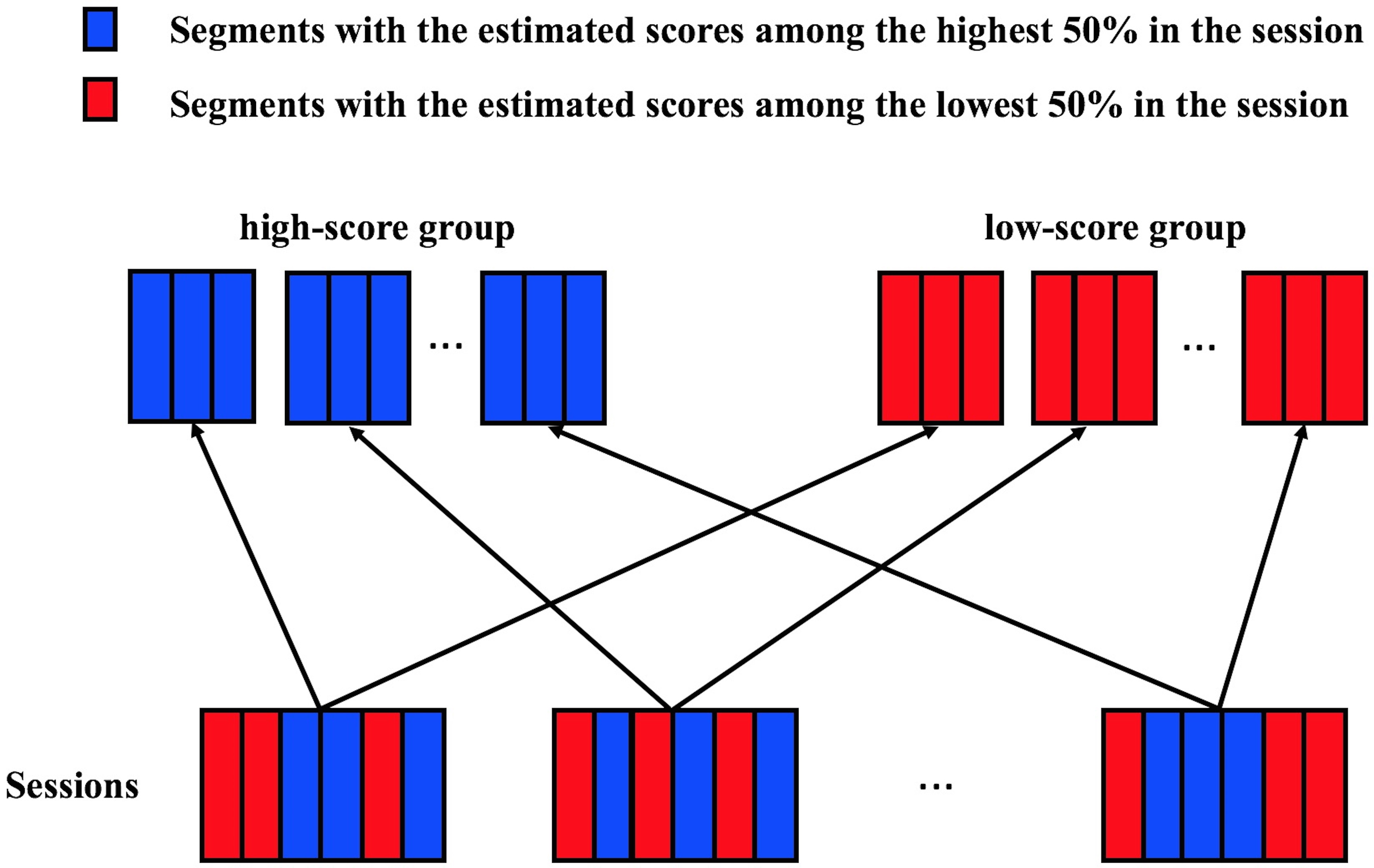}
  \vspace{-0.1cm}
  \caption{Divide the segments in terms of the relative performance within the session.}
  \label{fig:high-low}
   \vspace{-0.2cm}
\end{figure}

\begin{figure}[htb]
  \centering
  \includegraphics[width=12cm]{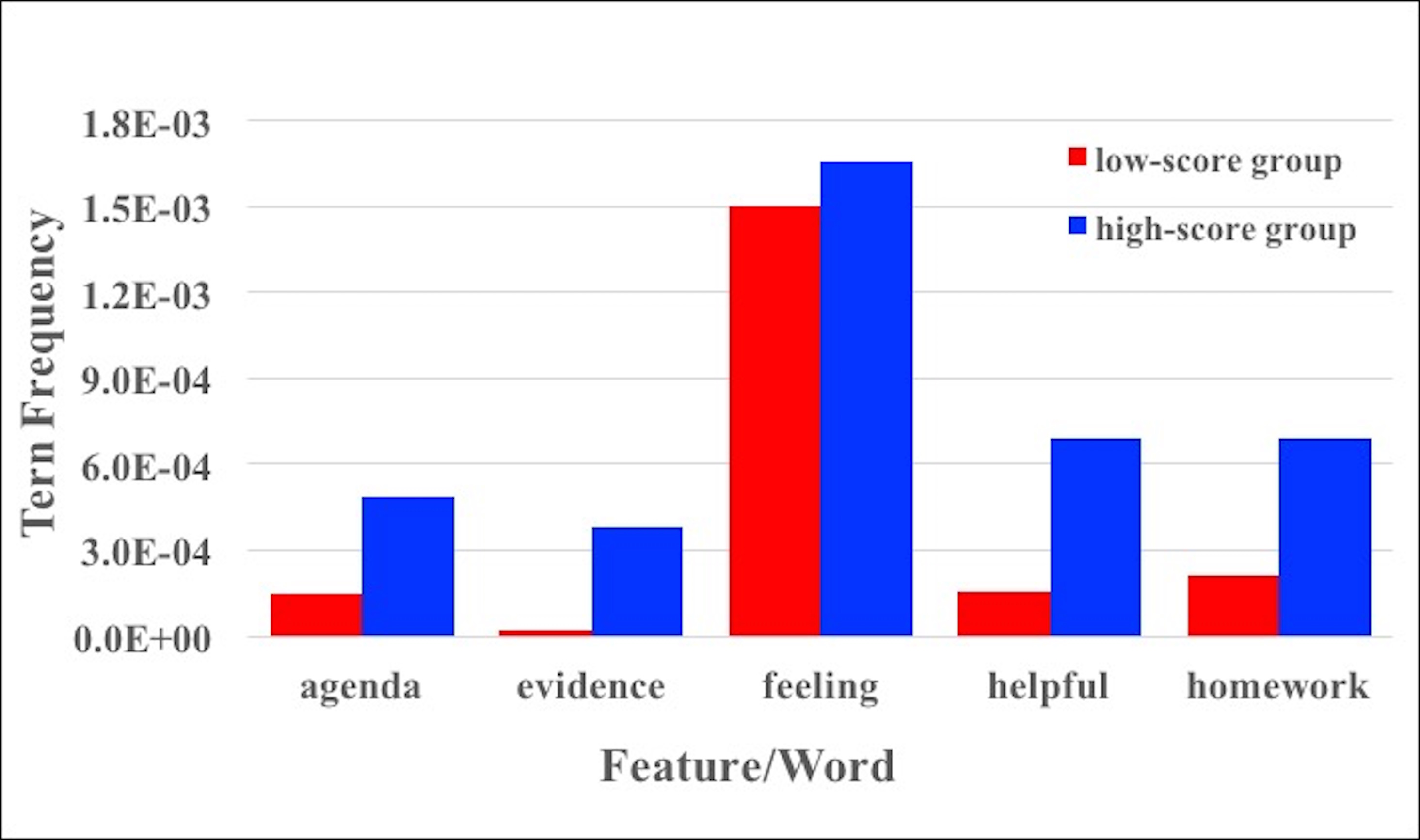}
  \vspace{-0.1cm}
  \caption{Comparison of term frequencies of key words between the low-score groups and the high-score groups; low-score group consists of the segments whose estimated scores are among the lowest 50\% in the session they belong to, high-score group  consists of the segments whose estimated scores are among the highest 50\% in the session they belong to.}
  \label{fig:words}
   \vspace{-0.2cm}
\end{figure}

We recorded the attention weights from the predictor using the best approach in Table~\ref{tab:result1} for a subset of the sessions in the test set, each of which consisted of ten segments (of $M=40$ utterances) to observe their behavior qualitatively through time. 
The average attention weights through time, across the selected sessions, for the CTRS codes \emph{agenda}, \emph{homework}, \emph{understand} and \emph{total score} are presented in Fig.~\ref{fig:attention}. As we can see, the attention mechanism assigns higher weights in the beginning when predicting the CTRS dimensions of \emph{agenda} and \emph{homework}. In CBT, the therapist sets the agenda collaboratively with the client to establish key topics to be discussed and reviews the homework in the early stages of a session. We also observe that the tail of the \emph{homework} curve goes up because the therapist is expected to assign new homework to their patient at the end of the session. \emph{Understanding} is the CTRS code used to evaluate the listening and empathic skills of the therapist and whether he/she successfully captured the patient's `internal reality' throughout a session. Thus, this therapist behavior is, on average, approximately equally important in each segment and, as a result, the attention weights are evenly distributed. In general, the average attention weights of the codes \emph{understanding} and \emph{total score} seem to be much more evenly distributed through time compared to those of \emph{agenda} and \emph{homework}, which partially explains that the \emph{uneven} mode is surpassed by the \emph{even} mode for predicting most of the CBT codes but achieves the best results for \emph{agenda} and \emph{homework}. The different behaviors between the attention weights distributions for different codes suggest that the \emph{even} mode and \emph{uneven} mode of our approach complement each other while evaluating the session quality from various perspectives. It is also interesting to point out that similar conclusions are drawn in our concurrent work in \citep{flemotomos2021automated}, despite the different approaches that are followed.

\subsection{Word distributions over segments}

Since we do not have reference (expert-provided) scores at the segment level, it is not possible to directly evaluate the accuracy of the segment quality estimates and confirm the fluctuations within a session. We can, however, perform a simple assessment by comparing the occurrence of the most informative words. To that end, we perform a backward selection to find the subset of the five best features/words for predicting the total CTRS scores using the tf-idf features of each session. The words selected are `agenda', `evidence', `feeling', `helpful' and `homework'. The Spearman correlations between the tf-idf features of these words and the total CTRS scores all fall into the range [0.6, 0.8]. These correlation scores indicate that the particular words tend to exist more frequently in sessions with higher total CTRS scores.

Again, by using the best approach in Table~\ref{tab:result1}, we obtain the segment quality estimates ($M=40$) of the sessions in the testing set. Fig.~\ref{fig:high-low} demonstrates that we cluster these segments into two groups containing: 1)  segments with estimated scores that are among the lowest 50\% in the session they belong to, denoted as ``low-score group"; 2) segments with estimated scores that are among the highest 50\% in the session they belong to, denoted as ``high-score group". We then compute the term frequency of the 5 words described above for both groups. As illustrated in Fig.~\ref{fig:words}, all of those words are more likely to exist in a segment with a high estimated score. For the words `agenda', `evidence', `helpful' and `homework', the term frequencies of the ``high-score group" are more than three times higher than the ones of the ``low-score group". These comparisons suggest that the estimated segment scores can provide insights into the variability in a therapist's performance within a session.

%
%
%

\section{Conclusion}

This paper introduces a hierarchical framework to evaluate the quality of psychotherapy conversations with a focus on cognitive behavioral therapy (CBT). We split sessions into blocks (conversation segments), employ BERT to learn segment embeddings, and use those features within an LSTM-based model to make predictions about session quality. We additionally implement a local quality estimator to model the estimated session quality as a linear combination of the segment-level quality estimates. The experimental results show a substantial gain over baselines. They suggest that incorporating such a local quality estimator leads to better segment representations and consistent improvements for assessing the overall session quality in most of the CTRS codes. In addition, we discuss how the estimated scores of the segment  
benefit the prediction tasks by comparing the differences of the segments within the same session. We should note that an important added benefit of our proposed approach is enhanced interpretability of the predicted results. By modeling the session quality as a function of local estimates, we get insights into specific salient parts of the therapy session and into how particular conversation segments contribute to the overall CTRS scores.

Our future work will focus on incorporating prosodic features into the hierarchical framework and on improved session segmentation approaches by detecting the topic shift. We will also explore whether utilizing the role information (therapist
vs. patient talking) can benefit the quality prediction. Finally, another area of interest is the potential generalization of our approach, so that it can be applied to different domains and therapeutic approaches; for example, for the prediction of global scores in Motivational Interviewing (MI) sessions.

\section*{Acknowledgments}

This work was funded by the National Institutes of Mental Health (NIMH), grant number R56 MH118550. UCC data collection was supported by the National Institute of Alcoholism and Alcohol Abuse (NIAAA), grant number R01 AA018673. CBT data collection was supported by the Philadelphia Department of Behavioral Health and Intellectual Disability Services (DBHIDS). We also appreciate the support and contribution of the University of Utah Counseling Center.



\bibliographystyle{elsarticle-harv} 
\bibliography{mybib}

\appendix
\section{Results for Different Segment Length}
\label{sec:M_v}

Table~\ref{tab:m_value} shows the evaluation results for different segment length using \emph{BERT-small + LSTM}, BERT fine-tuning is employed for all the tasks. The performance of the task is improved as we increase the number of $M$ from 1 to 40. However, when we set $M=80$, the number of words exceed the maximum sequence length (512) of BERT. The performance degrades because the long chunks cannot be fully fed into the model.

\begin{table}[htb]
\caption{Evaluation results for different segment length, $M$: \#utterances/segment}
  \label{tab:m_value}

\vspace{0.3cm}

\centering

\begin{tabular}{|c|c|c|c|}
\hline
M  & Average words per segment & RMSE/MAE           & F1 scores (\%) \\ \hline
1  & 8.4                       & 9.88/7.89          & 62.2           \\ \hline
5  & 41.8                      & 9.36/7.33          & 66.9           \\ \hline
20 & 166.2                     & 8.85/7.07          & 70.2           \\ \hline
40 & 327.9                     & \textbf{8.78/6.97} & \textbf{70.7}  \\ \hline
80 & 647.5                     & 9.02/7.14          & 69.5           \\ \hline
\end{tabular}
\end{table}


\end{document}